\RequirePackage[2020/02/08]{latexrelease}
\documentclass[final]{clv3}

\usepackage[dvipsnames]{xcolor}
\usepackage{amsmath}
\usepackage{dsfont}
\usepackage{enumitem}
\usepackage{hyperref}
\usepackage{tikz}
\usepackage{tikz-qtree}
\usetikzlibrary{decorations.pathreplacing,calc,tikzmark,calligraphy,positioning,arrows.meta,fit}
\usepackage[ruled,linesnumbered,algo2e]{algorithm2e}
\usepackage[defaultsans]{lato}
\usepackage{bm}
\usepackage{booktabs}
\usepackage{multirow}
\usepackage{colortbl}
\usepackage{float}
\usepackage[scale=0.9]{sourcecodepro}
\usepackage{appendix}

\definecolor{darkblue}{rgb}{0, 0, 0.5}
\definecolor{mdtRed}{rgb}{0.46, 0, 0}
\hypersetup{colorlinks=true,citecolor=mdtRed, linkcolor=mdtRed, urlcolor=mdtRed}

\makeatletter

\def\unwind@subpic#1{%
\edef\subpicid{#1}%
\ifx\subpicid\pgfpictureid
\else
\expandafter\ifx\csname pgf@sh@pi@#1\endcsname\relax
\pgf@xa=\pgf@x
\pgf@ya=\pgf@y
\pgfsys@getposition{\pgfpictureid}\pgf@shape@current@pos
\pgf@process{\pgfpointorigin\pgf@shape@current@pos}%
\advance\pgf@xa by-\pgf@x%
\advance\pgf@ya by-\pgf@y%
\pgf@process{\pgfpointorigin\subpic@parent@pos}%
\advance\pgf@xa by \pgf@x%
\advance\pgf@ya by \pgf@y%
\pgf@x=\pgf@xa
\pgf@y=\pgf@ya
\else
\pgfsys@getposition{\csname pgf@sh@pi@#1\endcsname}\subpic@parent@pos%
{%
  \pgfsettransform{\csname pgf@sh@nt@#1\endcsname}%
  \pgf@pos@transform{\pgf@x}{\pgf@y}%
  \global\pgf@x=\pgf@x
  \global\pgf@y=\pgf@y
}%
\unwind@subpic{\csname pgf@sh@pi@#1\endcsname}%
\fi
\fi
}

\def\pgf@shape@interpictureshift#1{%
\def\subpic@parent@pos{\pgfpointorigin}%
\unwind@subpic{\csname pgf@sh@pi@#1\endcsname}%
}

\makeatother

\newif\ifshowcomments
\showcommentsfalse
\newcommand{\abucomment}[1]{\ifshowcomments\textcolor{red}{[#1]}\fi}
\newcommand{\tomcomment}[1]{\ifshowcomments\textcolor{blue}{[#1]}\fi}

\newif\ifshowchanges
\showchangesfalse
\newcommand{\change}[1]{\ifshowchanges{\color{blue}#1}\else#1\fi}

\newcommand{\Equals}{\hspace{-0.2em}=\hspace{-0.2em}}

\issue{1}{1}{2016}

\runningtitle{A Probabilistic Generative Grammar for Semantic Parsing}

\runningauthor{Abulhair Saparov}

\begin{document}

\title{A Probabilistic Generative Grammar for Semantic Parsing}

\author{Abulhair Saparov}
\affil{Carnegie Mellon University}

\maketitle

\begin{abstract}
Domain-general semantic parsing is a long-standing goal in natural language processing, where the semantic parser is capable of robustly parsing sentences from domains outside of which it was trained. Current approaches largely rely on additional supervision from new domains in order to generalize to those domains. We present a generative model of natural language utterances and logical forms and demonstrate its application to semantic parsing. Our approach relies on domain-independent supervision to generalize to new domains. We derive and implement efficient algorithms for training, parsing, and sentence generation. The work relies on a novel application of hierarchical Dirichlet processes (HDPs) for structured prediction, which we also present in this manuscript.

This manuscript is an excerpt of chapter 4 from the Ph.D. thesis of \citet{SaparovThesis2022}, where the model plays a central role in a larger natural language understanding system.

This manuscript provides a new simplified and more complete presentation of the work first introduced in \citet{DBLP:conf/conll/SaparovSM17}. The description and proofs of correctness of the training algorithm, parsing algorithm, and sentence generation algorithm are much simplified in this new presentation. We also describe the novel application of hierarchical Dirichlet processes for structured prediction. In addition, we extend the earlier work with a new model of word morphology, which utilizes the comprehensive morphological data from Wiktionary.
\end{abstract}

\section{Introduction}

Accurate and efficient semantic parsing is a long-standing goal in natural language processing. Existing approaches are quite successful in particular domains \citep{DBLP:conf/uai/ZettlemoyerC05,DBLP:conf/emnlp/ZettlemoyerC07,DBLP:conf/acl/WongM07,DBLP:journals/coling/LiangJK13,DBLP:conf/emnlp/KwiatkowksiZGS10,DBLP:conf/emnlp/KwiatkowskiZGS11,DBLP:conf/emnlp/KwiatkowskiCAZ13,DBLP:conf/aaai/LiLS13,DBLP:conf/emnlp/WangKZ14,DBLP:conf/naacl/ZhaoH15,DBLP:conf/acl/DongL16,DBLP:conf/acl/RabinovichSK17}. However, they are largely domain-specific, relying on additional supervision such as a lexicon that provides the semantics or the type of each token in a set \citep{DBLP:conf/uai/ZettlemoyerC05,DBLP:conf/emnlp/ZettlemoyerC07,DBLP:conf/emnlp/KwiatkowksiZGS10,DBLP:conf/emnlp/KwiatkowskiZGS11,DBLP:journals/coling/LiangJK13,DBLP:conf/emnlp/WangKZ14,DBLP:conf/naacl/ZhaoH15,DBLP:conf/acl/DongL16,DBLP:conf/acl/RabinovichSK17}, or a set of initial synchronous context-free grammar rules \citep{DBLP:conf/acl/WongM07,DBLP:conf/aaai/LiLS13}. To apply the above systems to a new domain, additional supervision is necessary. When beginning to read text from a new domain, humans do not need to re-learn basic English grammar. Rather, they may encounter novel terminology. With this in mind, our approach is akin to that of \citep{DBLP:conf/emnlp/KwiatkowskiCAZ13} where we provide domain-independent supervision to help train a semantic parser. More specifically, our semantic parsing model restricts the rules that may be learned during training to a set that characterizes the general syntax of English. While we do not explicitly present and evaluate an open-domain semantic parser, we hope our work provides \emph{a step} in that direction.

Knowledge plays a critical role in natural language understanding. Even seemingly trivial sentences may have a large number of ambiguous interpretations. Consider the sentence \textit{``Ada started the machine with the GPU,''} for example. Without additional knowledge, such as the fact that ``machine'' can refer to computing devices that contain GPUs, or that computers generally contain devices such as GPUs, the reader cannot determine whether the GPU is part of the machine or if the GPU is a device that is used to start machines. Context is highly instrumental to quickly and unambiguously understand sentences.

In contrast to most semantic parsers, which are built on discriminative models, our model is fully generative: To generate a sentence, the logical form is first drawn from a prior. A grammar then recursively constructs a derivation tree top-down, probabilistically selecting production rules from distributions that depend on the logical form. The semantic prior distribution provides a straightforward way to incorporate background knowledge, such as information about the types of entities and predicates, or the context of the utterance. Additionally, our generative model presents a promising direction to \emph{jointly} learn to understand and generate natural language. Further, our parser can return \emph{partial} parses of sentences, which is useful for sentences that contain a small number of unseen words, such as definitions of new tokens. This can be exploited to learn new tokens and concepts outside of training.

In section \ref{sec:hdp_structured_prediction}, we present a novel application of hierarchical Dirichlet processes (HDPs) to structured prediction. We use this HDP model within our semantic parsing model, where HDPs are used to model dependence on logical forms. A mathematical description of the semantic parsing model is given in section \ref{sec:grammar}. In section \ref{sec:induction}, we describe the algorithms for training, parsing, and generation, including details on their implementation. In section \ref{section:semantic_parsing_experiments}, we apply this parsing approach to the \textsc{GeoQuery} and \textsc{Jobs} datasets \citep{DBLP:conf/aaai/ZelleM96, DBLP:conf/emnlp/TangM00}, using the Datalog representation of the provided logical form labels, and demonstrate that the accuracy of the parsed logical forms is comparable to that of the state-of-the-art on these datasets.

\section{Hierarchical Dirichlet processes for structured prediction} \label{sec:hdp_structured_prediction}

\tomcomment{Tom: The first mention of anything related to language is more than 10 pages into the chapter.  That makes the discussion up to section 4.2 very very abstract and ungrounded.} \abucomment{Abu: hmm, how would i rectify this? i want to keep the presentation of HDPs to be general since the novel application is also part of our contribution. maybe by adding a driving example that's not related to language?}

In order to describe our novel application of HDPs for structured prediction, which play a central role in our semantic parsing model, we must first define some notation as well as useful properties of Dirichlet processes.

The \emph{Dirichlet process} (DP) \citep{Ferguson73} is a distribution over probability distributions (i.e. samples from a DP are themselves distributions). If a distribution $G$ is drawn from a DP, we can write
\vspace{-1.0em}
\begin{equation}
    G\sim \text{DP}(\alpha,H),
\end{equation}
where the DP is characterized by two parameters: a concentration parameter $\alpha>0$ and a base distribution $H$. The DP has the useful property that $\mathbb{E}[G] = H$, and the concentration parameter $\alpha$ describes the ``closeness'' of $G$ to the base distribution $H$. If $\alpha$ is small, $G$ is more different from the base distribution $H$. If $\alpha$ is large, $G$ is more similar to $H$.

DPs are often used in statistical machine learning models where observations $y_1,y_2,\hdots$ are distributed according to $G$, such as in:
\begin{align}
    G &\sim \text{DP}(\alpha,H), \\
    y_1, y_2, \hdots &\sim G.
\end{align}
The joint probability of $y_1,\hdots,y_n$ is given by:
\begin{equation}
    p(y_1,\hdots,y_n) = \frac{\alpha^n\Gamma(\alpha)}{\Gamma(\alpha + n)} \prod_{k=1}^{m} H(y_k^*) \Gamma(n_k),
\end{equation}
where $y_k^*$ are the unique values of $y_1,\hdots,y_n$, $m$ is the number of such values, $n_k \triangleq \#\{i : y_i = y_k^*\}$ is the number of times $y_k^*$ appears in $y_1,\hdots,y_n$, and $\alpha^n\Gamma(\alpha)/\Gamma(\alpha + n)$ is the normalization term.

In these models, the \emph{Chinese restaurant process} (CRP) \citep{Aldous1985} provides a convenient equivalent description:
\begin{align}
    \phi_1,\phi_2,\hdots &\sim H, \\
    z_1 &= 1, \\
    z_{i+1} &=
		\begin{cases}
			k & \text{with probability } \frac{n_k}{\alpha + i}, \\
			k^{\text{new}} & \text{with probability } \frac{\alpha}{\alpha + i},
		\end{cases} \\
	y_i &= \phi_{z_i},
\end{align}
where $n_k \triangleq \#\{j \le i : z_j = k\}$ is the number of times $k$ appears in $\{z_1,\hdots,z_i\}$, $k^{\text{new}} \triangleq \max\{z_1,\dots,z_i\} + 1$ is the next integer that doesn't appear in $\{z_1,\hdots,z_i\}$. The analogy to a restaurant is to imagine a restaurant with a countably infinite sequence of tables, labeled $1, 2, 3, \hdots$ The first person comes into the restaurant and sits at table $1$. For each subsequent person that enters the restaurant, they choose to sit at a table with probability proportional to the number of people already sitting at that table. Otherwise, they choose to sit at an empty table with probability proportional to $\alpha$. $z_i$ indicates which table the $i^{\scriptsize th}$ customer chose to sit, $n_k$ is the number of people sitting at table $k$, and $k^{\text{new}}$ is the index of the next unoccupied table. Each table is assigned a sample from $H$, independently and identically distributed (i.i.d.), where $\phi_i$ is the sample assigned to table $i$. Each observation $y_i$ is the sample from $H$ that is assigned to the table that the $i^{\scriptsize th}$ customer chose to sit (i.e. table $z_i$). The CRP provides a simple algorithm to generate samples from a DP model. Notice that if $\alpha$ is very large, every customer is likely to choose to sit at a new table, and so each $y_i$ is likely to be drawn i.i.d. from $H$ (and therefore, $G$ would be very similar to $H$). The opposite is true in the case where $\alpha$ is small, where $G$ would be heavily concentrated on a small handful of observations, as each customer is more likely to sit at a table with existing customers. The CRP is \emph{exchangeable} which is useful property where the joint distribution of table assignments $\bm{z}$ is independent of their order. That is, for any permutation of the integers $\sigma$:
\begin{equation}
    p(z_1,z_2,\hdots) = p(z_{\sigma(1)}, z_{\sigma(2)}, \hdots).
\end{equation}

Note that this presentation of the DP differs from the classical presentation, where the DP is part of a mixture model, as in:
\begin{align}
    G &\sim \text{DP}(\alpha,H), \\
    \theta_1, \theta_2, \hdots &\sim G, \\
    y_i &\sim F(\theta_i),
\end{align}
where $F(\theta_i)$ is a distribution with parameter $\theta_i$. If $H$ is a conjugate prior of $F$, then an efficient Gibbs sampling algorithm is available, for example if $H$ is a Dirichlet distribution and $F$ is a multinomial, or if both $H$ and $F$ are normal distributions. In this manuscript, $F$ is assumed to be the delta function (the distribution whose samples are identical to the input parameter), and no assumptions are made on $H$ other than there exists an efficient way to compute the prior probability $p(\phi_i)$.

\subsection{Hierarchical Dirichlet processes} \label{sec:hdp}

The DP can be used as a component in larger models. The \emph{hierarchical Dirichlet process} (HDP) \citep{journals/jasa/Teh06} is a hierarchy of random variables, where each random variable is a distributed according to a Dirichlet process whose base distribution is given by the parent node in the hierarchy. Suppose each observation $y_i$ is coupled with a parameter $x_i$ that indicates the source node from which to sample the observation. Let the label of the root node in the hierarchy be $\textbf{0}$, and the model can be written:
\begin{align}
    G^{\textbf{n}} &\sim \begin{cases}
        \text{DP}(\alpha^{\textbf{0}},H) & \text{if } \textbf{n} = \textbf{0}, \\
        \text{DP}(\alpha^{\textbf{n}},G^{\text{parent}(\textbf{n})}) & \text{otherwise},
    \end{cases} \\
    y_i &\sim G^{x_i},
\end{align}
for all nodes in the hierarchy $\textbf{n}$. An equivalent ``Chinese restaurant'' representation may be written, which is coined a \emph{Chinese restaurant franchise} (CRF), where each node $\textbf{n}$ has a restaurant. For simplicity, assume that all $x_i$ are leaf nodes, then the CRF is written:
\begin{align}
    \phi_1,\phi_2,\hdots &\sim H, \\
    z_1^{\textbf{n}} &= 1, \\
    z_{i+1}^{\textbf{n}} &=
		\begin{cases}
			k & \text{with probability } \frac{n_k^{\textbf{n}}}{\alpha^{\textbf{n}} + i}, \\
			k^{\text{new}} & \text{with probability } \frac{\alpha^{\textbf{n}}}{\alpha^{\textbf{n}} + i},
		\end{cases} \\
	\psi_i^{\textbf{n}} &=
	    \begin{cases}
	        \phi_{z_i^{\textbf{0}}} & \text{if } \textbf{n} = \textbf{0}, \\
	        \psi_{z_i^{\textbf{n}}}^{\text{parent}(\textbf{n})} & \text{otherwise},
	    \end{cases} \\
	y_i &= \psi_{u_i + 1}^{x_i},
\end{align}
for all nodes in the hierarchy $\textbf{n}$, where $n_k^{\textbf{n}} \triangleq \#\{j \le i : z_j^{\textbf{n}} = k\}$ is the number of customers at node $\textbf{n}$ sitting at table $k$, $\smash{k^{\text{new}} \triangleq \max\{z_1^{\textbf{n}},\dots,z_i^{\textbf{n}}\} + 1}$ is the next available table at node $\textbf{n}$, and $u_i \triangleq \#\{j < i : x_j = x_i\}$ is the number of previous observations drawn from node $\textbf{n}$. In this extended metaphor, whenever a customer sits at a new table in the restaurant at node $\textbf{n}\ne\textbf{0}$, a ``new customer'' appears in the parent node $\text{parent}(\textbf{n})$ which corresponds to this table. The $\psi_i^{\textbf{n}}$ are the samples from $G^{\textbf{n}}$. Note that the above model is valid only when $x_i$ is a leaf node. If $x_i$ were a parent node, then the output samples $\psi_j^{x_i}$ are used by both the child nodes of $x_i$ as well as the observations $y_i$. In the restaurant metaphor, the customers at node $x_i$ not only come from its child nodes but also from the observations. In this case, the $\psi_j^{x_i}$ that are assigned to the observations come after those assigned to child nodes (the order does not actually matter thanks to exchangeability, so long as the samples\slash customers are partitioned between the two). More precisely, $y_i$ would be equal to $\psi_{c^{\textbf{n}} + u_i + 1}^{x_i}$ where $c^{\textbf{n}} = \max\{z_i^{\textbf{c}} : \textbf{c}\in\text{children}(\textbf{n})\}$ is the number of $\psi_j^{x_i}$ used by the child nodes of $\textbf{n}$ (i.e. the number of customers that come from the child nodes of $\textbf{n}$).

\subsection{Inferring the source node $x$} \label{sec:inferring_hdp_source_node}

Sections \ref{section:dp_gibbs} and \ref{section:hdp_gibbs} describes how to efficiently obtain posterior samples of $\bm{z}$ (and therefore, $\bm{\phi}$ and $\bm{\psi}$) using \emph{Markov chain Monte Carlo} (MCMC), given a set of observations $y_i$ and the corresponding nodes $x_i$ from which they were sampled. But now consider the case where the $\bm{x}$ are random variables, and we encounter a new observation $y^*$, but the source node $x^*$ (from which $y^*$ was sampled) is unknown, and we would like to infer it. That is, we would like to compute:
\begin{align}
    \hspace{-2.3em}\arg\max_{x^*} p(x^* | y^*, \bm{x}, \bm{y}) &= \arg\max_{x^*} p(x^*) \int p(y^* | x^*, \bm{z}) p(\bm{z} | \bm{x}, \bm{y}) d\bm{z}, \\
        &\approx \arg\max_{x^*} \frac{p(x^*)}{N_{\text{samples}}} \sum_{\bm{z}^{(t)} \sim \bm{z} \mid \bm{x}, \bm{y}} p(y^* \mid x^*, \bm{z}^{(t)}, \bm{\psi}^{(t)}, \bm{\phi}^{(t)}), \label{eq:source_node_objective} \\
    &\text{where } p(y^* \mid x^*, \bm{z}, \bm{\psi}, \bm{\phi}) = p(\psi_{\text{new}}^{x^*} \hspace{-0.2em}=\hspace{-0.1em} y^* \mid \bm{z}, \bm{\psi}, \bm{\phi}). \nonumber
\end{align}
This quantity is computed as in equations \ref{eq:new_table_probability_non_root} and \ref{eq:new_table_probability_root}. The $\arg\max$ over this objective is a discrete optimization problem, which, if solved naïvely, would require computing the objective function for every node $\textbf{n}$ in the tree. This is intractable if the tree is very large. Therefore, we present a branch-and-bound algorithm to perform this optimization efficiently.

\begin{algorithm2e}
\footnotesize
\let\oldnl\nl
\newcommand{\nonl}{\renewcommand{\nl}{\let\nl\oldnl}}
\SetNlSty{}{\color{RedOrange}\sffamily}{}
\SetAlgoBlockMarkers{}{}
\SetKwProg{Fn}{function}{}{}
\SetKwIF{If}{ElseIf}{Else}{if}{ }{else if}{else }{}
\SetKw{Continue}{continue}
\SetKwFunction{FBranchAndBound}{\small branch\_and\_bound}
\SetKwFor{ForEach}{for each}{do}{end}
\SetKwProg{uForEach}{for each}{ do}{}
\SetKwProg{Fn}{function}{}{}
\AlgoDisplayBlockMarkers\SetAlgoVlined
\SetAlCapNameFnt{\small}
\SetAlCapFnt{\small}
\SetNoFillComment
\DontPrintSemicolon
\SetInd{0.0em}{0.8em}
    \Fn{\FBranchAndBound{objective function $f$, heuristic $h$, domain $X$}}{
        $C$ is an empty list \;
        $Q$ is an empty priority queue \;
        $Q\texttt{.push}(X, \infty)$ \;
        \While{$Q$ not empty}{
            $(S,v) = Q\texttt{.pop}()$ \;
            \uIf{$S = \{x\}$ is a singleton}{
                $C\texttt{.add}(x, f(x))$ \; \label{line:branch_and_bound_singleton_case}
            }
            \Else{
                $(S_1,\hdots,S_n) = \texttt{branch}(S)$ \;
                \For{$i = 1,\hdots,n$}{
                    $Q\texttt{.push}(S_i, h(S_i))$ \;
                }
            }

            \tcc{check termination condition}
            \If{there are $k$ elements in $C$ with priority at least $v$}{
                break \;
            }
        }
        \Return{$C$} \tcc{the $k$ elements of $X$ that maximize $f$}
    }
    \caption{Pseudocode for a generic brand-and-bound algorithm for $k$-best discrete optimization.}
    \label{alg:branch_and_bound}
\end{algorithm2e}

\emph{Branch-and-bound} \citep{Land1960} is a method for solving discrete optimization problems. Pseudocode is shown in algorithm \ref{alg:branch_and_bound}. Given an objective function $f$, heuristic $h$, and search space $X$, the algorithm returns the $k$-best elements of $X$ that maximize the objective $f$. The algorithm requires that the heuristic $h$ be an \emph{upper bound} for $f$. That is, for any set $S$,
\begin{equation}
    h(S) \ge \max_{x\in S} f(x).
\end{equation}
The algorithm begins by considering the full search space $X$. A procedure called $\texttt{branch}$ then partitions $X$ into $n$ disjoint subsets $X_i$ (this procedure is specific to the optimization problem). Each subset is pushed onto the priority queue, with its key given by the heuristic $h(X_i)$. Then, for each iteration of the main loop, pop a set $S$ from the priority queue, and repeat the process: using $\texttt{branch}$, partition $S$ into $(S_1,\hdots,S_n)$, and then push each subset into the priority with key $h(S_i)$. If $S=\{x\}$ is a singleton set only containing the element $x$, then add it to a list of potential solutions. The algorithm terminates when there are $k$ potential solutions whose objective function values are at least the priority of $S$, or when the priority queue becomes empty. Once the algorithm terminates, the objective function values of the returned solutions are at least as large as the heuristic of the remainder of the search space. And since $h$ is an upper bound for $f$, the returned solutions are guaranteed to be optimal.

We develop a branch-and-bound algorithm to perform the optimization in equation \ref{eq:source_node_objective}. The HDP hierarchy provides a convenient search tree structure for the optimization. Let $D(\textbf{n})$ be the set of descendent nodes of $\textbf{n}$, including $\textbf{n}$ itself. The function $\texttt{branch}(D(\textbf{n}))$ is defined to partition $D(\textbf{n})$ into $(\{\textbf{n}\}, D(\textbf{c}_1), \hdots, D(\textbf{c}_n))$ where $\textbf{c}_i$ are the child nodes of $\textbf{n}$. We define a heuristic for $D(\textbf{n})$:
\begin{equation}
    \hspace{-0.5em}h(D(\textbf{n})) = \frac{h_x(D(\textbf{n}))}{N_{\text{samples}}} \hspace{-0.1em}\sum_{t=1}^{N_{\text{samples}}} \hspace{-0.2em}\max_{\{k:n_k^{\textbf{n}} > 0\}} \left\{\mathds{1}\{\psi_k^{\textbf{n}} = y^*\} , p(\psi_{\text{new}}^{\textbf{n}} = y^*) \right\}
\end{equation}
where $h_x(S)$ is an upper bound on the prior $h_x(S) \ge \max_{x\in S} p(x)$, the $\max$ is taken over all occupied tables in the restaurant at node $\textbf{n}$, and the references to $\psi$ within the sum are for the $t^{\scriptsize th}$ sample, $\psi^{(t)}$. $D(\textbf{n})$ can be sparsely represented in the implementation as a simple pointer to $\textbf{n}$. The heuristic is convenient since it can be computed only using the information available at node $\textbf{n}$, and so its running time is not a function of the size of the HDP hierarchy, as long as the heuristic on the prior $h_x(\cdot)$ is easy to compute. Furthermore, our algorithm avoids the recursion in the computation of $p(\psi_{\text{new}}^{\textbf{n}})$, since the term $p(\psi_{\text{new}}^{\text{parent}(\textbf{n}}))$ was already computed in the computation of the heuristic for the \emph{parent node}, and our algorithm re-uses it in future heuristic evaluations.

\newtheorem{thm}{Thm}
\begin{thm}
	The heuristic $h(D(\textbf{n}))$ is an upper bound on $\max_{x\in D(\textbf{n})} f(x)$ where $f$ is the objective function given by equation \ref{eq:source_node_objective}.
\end{thm}
\renewcommand\qedsymbol{}
\begin{proof}
	Consider any node $\textbf{m}\in D(\textbf{n})$ a descendant of $\textbf{n}$, and any MCMC sample $t$. We first aim to show that the quantity within the sum is an upper bound:
	\begin{equation}
        \max_{\{k:n_k^{\textbf{n}} > 0\}} \left\{\mathds{1}\{\psi_k^{\textbf{n}} = y^*\} , p(\psi_{\text{new}}^{\textbf{n}} = y^*) \right\} \ge p(y^* | x=\textbf{m}, \bm{z}^{(t)}, \bm{\psi}^{(t)}, \bm{\phi}^{(t)}).
	\end{equation}
	Since the right-hand side is equal to $p(\psi_{\text{new}}^{\textbf{m}} = y^*)$, the bound is trivially true in the case where $\textbf{m}=\textbf{n}$. So we can assume without loss of generality that $\textbf{m}\ne\textbf{n}$, and the right-hand side can be written:
	\begin{align}
        p(&y^* \mid x=\textbf{m}, \bm{z}^{(t)}, \bm{\psi}^{(t)}, \bm{\phi}^{(t)}) = p(\psi_{\text{new}}^{\textbf{m}} = y^*), \\
            &= \frac{\alpha^{\textbf{m}} p(\psi_{\text{new}}^{\text{parent}(\textbf{m})} \hspace{-0.4em}= y^*)}{\alpha^{\textbf{m}} + n^{\textbf{m}}} + \hspace{-1em}\sum_{\{k':n_{k'}^{\textbf{m}} > 0\}}\hspace{-1em} \frac{n^{\textbf{m}}_{k'} \mathds{1}\{\psi_{k'}^{\text{parent}(\textbf{m})} \hspace{-0.4em}= y^*\}}{\alpha^{\textbf{m}} + n^{\textbf{m}}},
	\end{align}
	according to equation \ref{eq:new_table_probability_non_root}. Since this expression is a convex combination of $\mathds{1}\{\psi_{k'}^{\text{parent}(\textbf{m})} \hspace{-0.4em}= y^*\}$ and $p(\psi_{\text{new}}^{\text{parent}(\textbf{m})} \hspace{-0.4em}= y^*)$, it is bounded above by:
	\begin{equation}
	    \le \max_{\{k' : n_{k'}^{\textbf{m}} > 0\}} \left\{\mathds{1}\{\psi_{k'}^{\text{parent}(\textbf{m})} \hspace{-0.4em}= y^*\}, p(\psi_{\text{new}}^{\text{parent}(\textbf{m})} \hspace{-0.4em}= y^*) \right\}.
	\end{equation}
	Due to equation \ref{eq:new_table_probability_non_root}, observe that $p(\psi_{\text{new}}^{\textbf{a}} \hspace{-0.4em}= y^*) \le p(\psi_{\text{new}}^{\text{parent}(\textbf{a})} \hspace{-0.4em}= y^*)$ for any node $\textbf{a}$. In addition, by construction of the HDP, the $\psi_k^{\textbf{a}}$ at any node $\textbf{a}$ are a subset of the $\smash{\psi_k^{\text{parent}(\textbf{a})}}$. That is, for all $k$, there is a $k'$ such that $\smash{\psi_k^{\textbf{a}} = \psi_{k'}^{\text{parent}(\textbf{a})}}$. These observations extend to all ancestors of $\textbf{a}$. Applying these two observations to the node $\textbf{m}$, we can conclude that the above expression is further bounded above by:
	\begin{equation}
	    \le \max_{\{k' : n_{k'}^{\textbf{n}} > 0\}} \left\{\mathds{1}\{\psi_{k'}^{\textbf{n}} \hspace{-0.4em}= y^*\}, p(\psi_{\text{new}}^{\textbf{n}} \hspace{-0.4em}= y^*) \right\}.
	\end{equation}
	We have shown that the quantity within the sum of the heuristic is an upper bound. Since by definition, $h_x(D(\textbf{n})) \ge \max_{x\in D(\textbf{n})} p(x) \ge p(x^* \Equals \textbf{m})$, the full heuristic $h(D(\textbf{n}))$ is an upper bound on $f(\textbf{m})$, the objective function evaluated at $\textbf{m}$, for all $\textbf{m}\in D(\textbf{n})$. Therefore, $h(D(\textbf{n})) \ge \max_{\textbf{m}\in D(\textbf{n})} f(\textbf{m})$. $\blacksquare$
\end{proof}

The branch-and-bound algorithm starts with the input set $D(\textbf{0})$, which is the set of all nodes in the tree, and will efficiently compute the $k$ most probable values of the source node $x^*$, from which the observation $y^*$ was sampled. Note that the above algorithm is easily extended to the case where the HDP is part of a mixture model (i.e. $F$ is not a delta function). To do so, replace each instance of $\mathds{1}\{\psi_k^{\textbf{n}} = y^*\}$ with $p(y^* \mid y^* \sim F(\theta_k^{\textbf{n}}), \bm{z}, \bm{\phi})$, for all $\textbf{n}$ and $k$.

The above algorithm can be generalized to the case where $x^*$ is restricted to a subset of the nodes $X$ in the hierarchy: $\arg\max_{x^*} p(x^* \mid x^*\in X, y^*, \bm{x}, \bm{y})$. In this case, the algorithm is started with the input set $D(\textbf{0})\cap X$. The $\texttt{branch}$ function is modified: $\texttt{branch}(D(\textbf{n}) \cap X)$ partitions the set $D(\textbf{n}) \cap X$ into $(\{\textbf{n}\} \cap X, D(\textbf{c}_1) \cap X, \hdots, D(\textbf{c}_n) \cap X)$ where $\textbf{c}_i$ are the child nodes of $\textbf{n}$.

\subsection{Infinite hierarchies}

To apply the HDP in our semantic parsing model, we need to be able to handle the case where the HDP hierarchy is infinite (but with finite height). That is, every non-leaf node in the hierarchy may have an infinite number of children. But this makes no difference in the MCMC algorithm to infer $\bm{z}, \bm{\phi}, \bm{\psi}$, since the number of given observations $(\bm{x}, \bm{y})$ is finite. We only need to compute and keep track of the variables that are associated with an observation (either at the current node or a descendant). Thus, the only nodes of the tree that we need to explicitly keep in memory are those of $\bm{x}$ and their ancestors, as the restaurants at all other nodes are empty. The explicitly-stored tree size is bounded by the product of the number of distinct $x_i$ and the height of the tree.

However, the branch-and-bound algorithm to find the most probable source node $x^*$ needs to be adapted, since the $\texttt{branch}$ function would otherwise return an infinite number of subsets. Consider any node $\bm{n}$ that has no observations (i.e. has an empty restaurant). Then by equation \ref{eq:new_table_probability_non_root}, $p(\psi_{\text{new}}^{\textbf{n}}) = \smash{p(\psi_{\text{new}}^{\text{parent}(\textbf{n})})} = \hdots = p(\psi_{\text{new}}^{\textbf{a}})$ where $\textbf{a}$ is the most recent non-empty ancestor of $\textbf{n}$. For such nodes, the objective function in equation \ref{eq:source_node_objective} can be simplified
\begin{equation}
    \frac{p(\textbf{n})}{N_{\text{samples}}} \sum_{\bm{z}^{(t)} \sim \bm{z} \mid \bm{x}, \bm{y}} p(\psi_{\text{new}}^{\textbf{a}} = y^*).
\end{equation}
Aside from the prior term $p(\textbf{n})$, all empty descendant nodes of $\textbf{a}$ have the same objective function value, which is independent of $\textbf{n}$. So to adapt the algorithm to the infinite hierarchy case, the $\texttt{branch}$ function is modified:
\begin{equation}
    \texttt{branch}(D(\textbf{a})) \hspace{0.2em}\textit{ returns } \bigg(\hspace{-0.1em} \{\textbf{a}\}, D(\textbf{c}_1), \hdots, D(\textbf{c}_n), \hspace{-0.1em}\bigcup_{i=n+1}^\infty \hspace{-0.2em} D(\textbf{c}_i) \hspace{-0.1em}\bigg),
\end{equation}
where $(\textbf{c}_1,\hdots,\textbf{c}_n)$ are the non-empty child nodes of $\textbf{a}$, and $(\textbf{c}_{n+1}, \textbf{c}_{n+2},$ $\hdots)$ are the empty child nodes of $\textbf{a}$. Next, in algorithm \ref{alg:branch_and_bound}, following line \ref{line:branch_and_bound_singleton_case}, we add a new else-if statement to check for the case that $S$ is a set of empty nodes. If so, $S$ is added to $C$, and we don't continue the search in the empty descendant nodes. The resulting adapted branch-and-bound algorithm correctly and efficiently solves the optimization problem for infinite hierarchies.

\subsection{Modeling dependence on discrete structures} \label{sec:hdp_structured_model}

HDPs can be used to learn distributions that depend on sequences of non-negative integers. Consider the data $\{(x_1,y_1),\hdots,$ $(x_n,y_n)\}$ where each $x_i\in\mathbb{Z}_{+}^h$ is a sequence of $h$ non-negative integers.\tomcomment{Tom: Where is $\mathbb{Z}_{+}^h$ defined?}\abucomment{Abu: this is just the standard notation for $h$-dimensional space of non-negative integers, but maybe i should avoid this jargon?} The distribution of $y_i$ is dependent on the value of $x_i$. We can use the HDP to learn the relationship of this dependence: construct a hierarchy of height $h$, where each non-leaf node has a countably infinite number of children, every child node corresponding to a non-negative integer. Here, each $x_i$ uniquely identifies a leaf node in the hierarchy by characterizing a path from the root $\textbf{0}$ to a leaf: the first integer in the sequence identifies the child of the root node, the second integer identifies the grandchild, and so on. The $y_i$ are then sampled from the corresponding leaf node. We can apply MCMC to learn the distributions of the $y_i$, and how those distributions relate to the integer sequences $x_i$.

Given a new observation $y^*$, the branch-and-bound algorithm can be used to find the most probable corresponding integer sequence $x^*$, but we need to be able to convert the output of the branch-and-bound into the corresponding integer sequence. The algorithm will output a list of the $k$ most probable source nodes from which $y^*$ is sampled, \emph{or} sets of empty source nodes (since the HDP hierarchy is infinite). More precisely, let $(o_1, \hdots, o_k)$ be the output of the branch-and-bound algorithm. For each $o_j$, there are two cases:
\begin{enumerate}
    \item $o_j$ is a single leaf node, in which case it is straightforward to convert the node into its corresponding integer sequence.
    \item $o_j$ is the set of empty descendants of a node $\textbf{a}$. In this latter case, it can be converted into an ``incomplete'' sequence of integers, where the first $L(\textbf{a})$ numbers of the sequence correspond to the node $\textbf{a}$, where $L(\textbf{a})$ is the level of $\textbf{a}$. This incomplete sequence represents the set of all integer sequences that begin with the same $L(\textbf{a})$ integers, that do not already explicitly exist in the tree.
\end{enumerate}
For example, let $\textbf{n}_a$ be the $a^{\scriptsize th}$ child of the root node $\textbf{0}$ in the HDP hierarchy. Let $\textbf{n}_{a,b}$ be the $b^{\scriptsize th}$ child of $\textbf{n}_a$, and so on. Suppose the training set contains only the sequences $(4,3,1)$, $(4,7,4)$, and $(4,8,2)$. Therefore, the nodes in the HDP with non-empty restaurants are: $\textbf{n}_{4,3,1}$, $\textbf{n}_{4,7,4}$, $\textbf{n}_{4,8,2}$, $\textbf{n}_{4,3}$, $\textbf{n}_{4,7}$, $\textbf{n}_{4,8}$, $\textbf{n}_{4}$, and $\textbf{0}$. If the branch-and-bound algorithm returns $\textbf{n}_{4,7,4}$, the corresponding output integer sequence is $(4,7,4)$. If instead, branch-and-bound returns the set of the empty descendant nodes of $\textbf{n}_{4}$, the corresponding output integer sequence is $(4,*\setminus\{3,7,8\},*)$. The `$*$' is a ``wildcard'' symbol that represents the set of all non-negative integers. Thus, $(4,*\setminus\{3,7,8\},*)$ represents the set of all integer sequences that start with $(4,\hdots)$ but do not start with $(4,3,\hdots)$, $(4,7,\hdots)$, or $(4,8,\hdots)$.

This model can be extended to the case where the $x_i \in \mathcal{X}$ have richer structure (e.g. $\mathcal{X}$ is the set of labeled trees, graphs, logical forms, etc)\change{, i.e. \emph{structured prediction}}. To do so, define $d$ functions $f_k:\mathcal{X}\to\mathbb{Z}_+$ that characterize an aspect of the input structures $x_i$. We call these functions $f_k$ \emph{feature functions}. For example, if $x$ is a labeled binary tree, $f_1(x)$ returns the label of the root node, and $f_2(x)$ returns the label of the left child, etc. The functions serve to map the structures $x_i$ into sequences of non-negative integers: $(f_1(x_i), \hdots, f_d(x_i))$. Then the above HDP model can be directly used to learn the relationship between these integer sequences and the distribution of the observations $y_i$. For a new observation $y^*$, the branch-and-bound algorithm will return the $k$ most likely integer sequences (possibly with wildcard symbols) that represent the unknown structure $x^*$. To convert the integer sequence $(w_1,\hdots,w_d)$ into the corresponding structure in $\mathcal{X}$, we can compute:
\begin{equation}
    f_1^{-1}(w_1) \bigcap \hdots \bigcap f_d^{-1}(w_d) \text{ where } f_k^{-1}(w_k) \triangleq \{x:f_k(x)\in w_k\}.
\end{equation}
Our code implements three functions to perform the above mapping between integer sequences and more structured representations in $\mathcal{X}$:
\begin{enumerate}
    \item \texttt{get\_feature($f$, $X$)}: Given a feature function $f$ and a set $X\subseteq\mathcal{X}$, return $\{f(x) : x\in X \}$.
    \item \texttt{set\_feature($f$, $X^{\text{old}}$, $w$)}: Given a feature function $f$, a set $X^{\text{old}}\subseteq\mathcal{X}$, and a non-negative integer $w\in\mathbb{Z}_+$, return $X^{\text{old}} \cap f^{-1}(w)$. This function is used in the case that $w_k$ is an integer (not a wildcard).
    \item \texttt{exclude\_features($f$, $X^{\text{old}}$, $W$)}: Given a feature function $f$, a set $X^{\text{old}}\subseteq\mathcal{X}$, and a finite set of non-negative integers $W\in\mathbb{Z}_+^*$, return $X^{\text{old}} \setminus f^{-1}(W)$. This is used in the case that $w_k$ is a wildcard $*\setminus W$.
\end{enumerate}

\change{
\subsection{Related work}

The HDP hierarchy in our proposed model in section \ref{sec:hdp_structured_model} resembles a decision tree \citep{DBLP:books/daglib/0023820}. The input features determine the path within the tree, and the output is sampled from a leaf node. \citet{DBLP:conf/acl/Teh06} constructs a language model using a \emph{hierarchical Pitman-Yor process} (HPY), which is a generalization of the HDP that exhibits power-law behavior. In their model, the HPY describes the distribution of the next character in a sequence of characters, conditioned on the previous $d$ characters. The sequence of preceding $d$ characters corresponds to the path in the hierarchy of depth $d$. Our approach is a novel application of HDPs for structured prediction, where the path in the hierarchy is a random variable which corresponds to the structure we aim to predict. Since the HDP hierarchies are infinite, the model does not a priori impose a limit on the number of possible structures or logical forms. An idea for future work is to replace the HDP in our model with the HPY to better capture power-law behavior which is prevalent in natural language.
}

\begin{figure}[h]
	\centering
	\begingroup
	\begin{align*}
		\textsf{S} &\to \textsf{N\hspace{0.4mm}:\hspace{0.4mm}\texttt{\color{RoyalBlue}select\_arg1} \hspace{0.1mm} VP\hspace{0.4mm}:\hspace{0.4mm}\texttt{\color{RoyalBlue}delete\_arg1}} \span\span \\[-0.3em]
		\textsf{VP} &\to \textsf{V\hspace{0.4mm}:\hspace{0.4mm}\texttt{\color{RoyalBlue}identity} \hspace{0.1mm} N\hspace{0.4mm}:\hspace{0.4mm}\texttt{\color{RoyalBlue}select\_arg2}} \span\span \\[-0.3em]
		\textsf{VP} &\to \textsf{V\hspace{0.4mm}:\hspace{0.4mm}\texttt{\color{RoyalBlue}identity}} \\[-0.3em]
		\textsf{N} &\to \text{``New Jersey''} 	&\hspace{4mm}\textsf{V} &\to \text{``borders''} \\[-0.3em]
		\textsf{N} &\to \text{``NJ''}				&\hspace{4mm}\textsf{V} &\to \text{``bordered''} \\[-0.3em]
		\textsf{N} &\to \text{``Pennsylvania''}	&\hspace{4mm}\textsf{V} &\to \text{``has''} \\[-0.3em]
		\textsf{N} &\to \text{``Michael Phelps''}	&\hspace{4mm}\textsf{V} &\to \text{``swims''} \\[-0.3em]
		\textsf{N} &\to \text{``tennis''}	&\hspace{4mm}\textsf{V} &\to \text{``plays''}
	\end{align*}
	\endgroup
	\caption{Example of a grammar in our framework. This example grammar operates on logical forms of the form \textit{predicate(first argument, second argument)}. The semantic function {\normalfont\color{RoyalBlue}\texttt{select\_arg1}} returns the first argument of the logical form. Likewise, the function {\normalfont\color{RoyalBlue}\texttt{select\_arg2}} returns the second argument. The function {\normalfont\color{RoyalBlue}\texttt{delete\_arg1}} removes the first argument, and {\normalfont\color{RoyalBlue}\texttt{identity}} returns the logical form with no change. In our work, the interior production rules (the first three listed above) are examples of rules that we specify, whereas the terminal rules and the posterior probabilities of \emph{all} rules are learned via grammar induction. A simplified semantic representation is shown here for the sake of illustration. Our semantic parser uses a richer semantic representation. Section \ref{sec:selecting_production_rules} provides more detail.}
	\label{fig:example_grammar}
\end{figure}

\section{Model: semantic grammar} \label{sec:grammar}

A grammar in our formalism operates over a set of nonterminals $\mathcal{N}$ and a set of terminal symbols $\mathcal{W}$. It can be understood as an extension of a context-free grammar (CFG) \citep{DBLP:journals/tit/Chomsky56} where the generative process for the syntax is dependent on a logical form, thereby coupling syntax with semantics. In the top-down generative process of a derivation tree, a logical form guides the selection of production rules. Production rules in our grammar have the form $A \to B_1\hspace{-1mm}\nobreak:\nobreak\hspace{-1mm}f_1 \hdots B_k\hspace{-1mm}\nobreak:\nobreak\hspace{-1mm}f_k$ where $A\in\mathcal{N}$ is a nonterminal, $B_i\in\mathcal{N}\cup\mathcal{W}$ are right-hand side symbols, and $f_i$ are \emph{semantic transformation functions}. These functions describe how to ``decompose'' this logical form when recursively generating the subtrees rooted at each $B_i$. Thus, they enable semantic compositionality. An example of a grammar in this framework is shown in figure \ref{fig:example_grammar}, and a derivation tree is shown in figure \ref{fig:example_derivation}. Let $\bm{\mathcal{R}}$ be the set of production rules in the grammar and $\bm{\mathcal{R}}_A$ be the set of production rules with left-hand nonterminal symbol $A$.

\begin{figure}
	\definecolor{lightgrey}{rgb}{0.95,0.95,0.95}
	\centering
	\tikzstyle{nonterminal}=[circle,draw,fill=lightgrey,inner sep=0,minimum size=7mm,thick]
	\begin{tikzpicture}[
		level 1/.style={sibling distance=2.4cm,level distance=8mm,line width=1pt,edge from parent path={(\tikzparentnode) -- (\tikzchildnode)}},
		level 2/.style={sibling distance=1.35cm,level distance=8mm,line width=1pt,edge from parent path={(\tikzparentnode) -- (\tikzchildnode)}}]
	\node[nonterminal,label=right:\small{\color{RoyalBlue}\texttt{borders(pa,nj)}}] (S) {\textsf{\textbf{S}}}
		child {node[nonterminal,label=left:\small{\color{RoyalBlue}\texttt{pa}}] (N) {\textsf{\textbf{N}}}
		child {node[text depth=0mm] (pa) {\textrm{``Pennsylvania''}}}
		}
		child {node[nonterminal,label=right:\small{\color{RoyalBlue}\texttt{borders(,nj)}}] (VP) {\textsf{\textbf{VP}}}
			child {node[nonterminal] (V) {\textsf{\textbf{V}}}
				child {node[text height=2mm,text depth=0mm] (borders) {\textrm{``borders''}}}
			}
			child {node[nonterminal,label=right:\small{\color{RoyalBlue}\texttt{nj}}] (Nobj) {\textsf{\textbf{N}}}
				child {node[text height=2mm,text depth=0mm] (nj) {\textrm{``NJ''}}}
			}
		};
	\end{tikzpicture}
	\caption{Example of a derivation tree under the grammar given in figure \ref{fig:example_grammar}. The logical form corresponding to every node is shown in blue beside the respective node. The logical form for V is {\normalfont\color{RoyalBlue}\texttt{borders(,nj)}} and is omitted to reduce clutter.}
	\label{fig:example_derivation}
\end{figure}

\subsection{Generative process} \label{sec:language_module_generative_process}

A \emph{derivation tree} in this formalism is a tree where every interior node is labeled with a nonterminal symbol in $\mathcal{N}$, every leaf is labeled with a terminal in $\mathcal{W}$, and the root node is labeled with the root nonterminal $S$. Moreover, every node in the tree is associated with a logical form: let $x^{\textbf{n}}$ be the logical form assigned to the tree node $\textbf{n}$, and $x^{\textbf{0}} = x$ for the root node $\textbf{0}$.

The generative process to build a derivation tree begins with the root nonterminal $S$ and a logical form $x$. The logical form $x$ is drawn from a prior distribution on logical forms $p(x)$. The generative process \emph{expands} $S$ by randomly drawing a production rule from $\bm{\mathcal{R}}_S$, \emph{conditioned} on the logical form $x$. This provides the first level of child nodes in the derivation tree. For example, if the rule $\textsf{S} \to B_1\hspace{-1mm}\nobreak:\nobreak\hspace{-1mm}f_1 \hdots B_k\hspace{-1mm}\nobreak:\nobreak\hspace{-1mm}f_k$ were drawn, the root node would have $k$ child nodes, $\textbf{n}_1, \hdots, \textbf{n}_k$, respectively labeled with the symbols $B_1, \hdots, B_k$. The logical form associated with each node is determined by the semantic transformation function: $x^{\textbf{n}_i} = f_i(x^{\textbf{0}})$. These functions describe the relationship between the logical form at a child node and that of its parent node. This process repeats recursively with every right-hand side nonterminal symbol, until there are no unexpanded nonterminal nodes. The sentence is obtained by taking the \emph{yield} (i.e. the concatenation) of the terminals in the tree.

The semantic transformation functions are specific to the semantic formalism and may be defined as appropriate to the application. In our semantic parsing experiments in section \ref{section:semantic_parsing_experiments}, we define a domain-independent set of transformation functions specific to the Datalog representation of \textsc{GeoQuery} and \textsc{Jobs} (e.g., one function selects the left $n$ conjuncts in a conjunction, another selects the $n^{\footnotesize th}$ argument of a predicate instance, etc). Some examples of these transformation functions are:
\begin{itemize}
    \item The function \texttt{\color{RoyalBlue}select\_left} returns the left conjunct of a conjunction. For example, given the Datalog expression \texttt{\color{RoyalBlue}(river(A),} \texttt{\color{RoyalBlue}loc(A,B),const(B,stateid(colorado)))}, this function returns \texttt{\color{RoyalBlue}river(A)}.
    \item The function \texttt{\color{RoyalBlue}delete\_left} returns a conjunction where the first conjunct is removed. For example, given \texttt{\color{RoyalBlue}(river(A),loc(A,B),} \texttt{\color{RoyalBlue}const(B,stateid(colorado)))}, this function returns \texttt{\color{RoyalBlue}(loc(A,B),} \texttt{\color{RoyalBlue}const(B,stateid(colorado)))}.
    \item The function \texttt{\color{RoyalBlue}select\_arg2} returns the second argument in an atomic formula. For example, given \texttt{\color{RoyalBlue}const(A,stateid(maine))}, this function returns \texttt{\color{RoyalBlue}stateid(maine)}.
\end{itemize}

Semantic transformation functions are allowed to fail, which is useful in defining richer transformation functions and providing more flexibility when designing the production rules of the grammar. If in the generative process, a transformation function returns failure, the generative process is restarted from the root (all progress up to the failure is discarded). As an example, failure enables the definition of transformation functions that check whether the input logical form satisfies a specific property: \texttt{\color{RoyalBlue}require\_binary\_conjunction} returns the input logical form, unchanged, if it is a conjunction of length 2; otherwise, it returns failure. Since failure can cause the generative process to repeatedly restart, the process of sampling using the generative process can be expensive. However, our approach does not generate sentences using this algorithm,\tomcomment{Then why are we discussing it here?}\abucomment{Abu: i could add ``The generative process is meant to help specify well-defined prior and conditional distributions of the random variables in the model.'' it has zero bearing on the actual computation that we do during inference; i could also add something early in chapter 2 to make this a more general statement about generative modeling} and as we show in section \ref{sec:parser_inference}, the performance of the parser is not adversely affected.

\subsection{Selecting production rules} \label{sec:selecting_production_rules}

The above description does not specify the conditional distribution from which rules are selected from $\bm{\mathcal{R}}_A$ given the logical form. There are many modeling options available in choosing this distribution, but we need a distribution that captures complex dependencies between the logical form and selected production rule. For example, consider the grammar in figure \ref{fig:example_grammar} and the logical form ${\color{RoyalBlue}\texttt{plays\_sport(}}$ ${\color{RoyalBlue}\texttt{michael\_phelps,tennis)}}$. When generating a sentence for this logical form, at the root nonterminal $\textsf{S}$, there is only one production rule available, $\textsf{S} \to \textsf{N\hspace{0.4mm}:\hspace{0.4mm}\texttt{\color{RoyalBlue}select\_arg1} \hspace{0.1mm} VP\hspace{0.4mm}:\hspace{0.4mm}\texttt{\color{RoyalBlue}delete\_arg1}}$, so this rule is selected. Now consider the child node corresponding to the nonterminal $\textsf{VP}$. Its logical form is ${\color{RoyalBlue}\texttt{plays\_sport(,tennis)}}$, which is the output of the function $\texttt{\color{RoyalBlue}delete\_arg1}$ when applied to the logical form at the root node. At this point, there are two choices of production rules with $\textsf{VP}$ on the left-hand side: $\textsf{VP} \to \textsf{V}\hspace{2mm}\textsf{N}$ and $\textsf{VP} \to \textsf{V}$. In this case, we want the conditional distribution to select $\textsf{VP} \to \textsf{V}\hspace{2mm}\textsf{N}$, since the most likely sentence that conveys the semantics in the logical form is ``plays tennis.'' In fact, $\textsf{VP} \to \textsf{V}\hspace{2mm}\textsf{N}$ should be selected even in when the logical form is ${\color{RoyalBlue}\texttt{plays\_sport(,baseball)}}$ or ${\color{RoyalBlue}\texttt{plays\_sport(,soccer)}}$ or almost any other sport. However, if the logical form were ${\color{RoyalBlue}\texttt{plays\_sport(,swimming)}}$, we want the conditional distribution to give higher probability to $\textsf{VP} \to \textsf{V}$, since the verb phrase ``swims'' is much more likely. Therefore, a desirable property of the conditional distribution for $\textsf{VP}$ production rules is that the distribution depends primarily on the predicate symbol (e.g. ${\color{RoyalBlue}\texttt{plays\_sport}}$) but also depends secondarily on the object argument (e.g. ${\color{RoyalBlue}\texttt{swimming}}$ or ${\color{RoyalBlue}\texttt{tennis}}$).

Our model uses the HDP to capture this dependence, as presented in section \ref{sec:hdp_structured_model}. Every nonterminal in our grammar $A\in\mathcal{N}$ will be associated with an HDP hierarchy. For each nonterminal, we specify a sequence of \emph{semantic feature functions}, $\{g_1,\hdots,g_m\}$, each of which, when given input logical form $x$, returns a non-negative integer. The HDP hierarchy is a complete infinite tree of height $m$, where every parent node has an infinite number of child nodes, one for each non-negative integer. The base distribution $H$ at the root of the HDP is over $\bm{\mathcal{R}}_A$.


Take, for example, the derivation in figure \ref{fig:example_derivation}. In the generative process where the node $\textsf{VP}$ is expanded, the production rule is drawn from the HDP associated with the nonterminal $\textsf{VP}$. Suppose the HDP was constructed using a sequence of two semantic features: $({\color{RoyalBlue}\texttt{predicate}}, {\color{RoyalBlue}\texttt{arg2}})$. In the example, the feature functions are evaluated with the logical form $\color{RoyalBlue}\texttt{borders(,nj)}$ and they return a sequence of two integers, the first is the identifier for the symbol {\color{RoyalBlue}\texttt{borders}} and the second is the identifier for the symbol {\color{RoyalBlue}\texttt{nj}}. This sequence uniquely identifies a path in the HDP hierarchy from the root node $\textbf{0}$ to a leaf node $\textbf{n}$. The production rule $\textsf{VP} \to \textsf{V}\hspace{2mm}\textsf{N}$ is drawn from this leaf node $G^{\textbf{n}}$, and the generative process continues recursively. As desired, the distribution of the selected production rule $G^{\textbf{n}}$ depends on the HDP source node $\textbf{n}$, which itself depends primarily on the first feature and secondarily on the second feature and so on (in this example, the ${\color{RoyalBlue}\texttt{predicate}}$ and ${\color{RoyalBlue}\texttt{arg2}}$ of the logical form are the first and second features, respectively).

In our implementation, the set of nonterminals $\mathcal{N}$ is divided into two disjoint groups: (1) the set of ``interior'' nonterminals, and (2) preterminals. The production rules of preterminals are restricted such that the right-hand side contains only terminal symbols. The rules of interior nonterminals are restricted such that only nonterminal symbols appear on the right side.
\begin{enumerate}
  \item For \textbf{preterminals}, $H$ is a distribution over sequences of terminal symbols. The sequence of terminal symbols is distributed as follows: first sample the length of the terminal from a geometric distribution (i.e. the number of words) and then generate each word in the sequence i.i.d. from a uniform distribution over a finite set of (initially unknown) terminals. Note that we do not specify a set of domain-specific terminal symbols in defining this distribution.
  \item For \textbf{interior nonterminals}, $H$ is a discrete distribution over a domain-independent set of production rules, which we specify. Since the production rules contain transformation functions, they are specific to the semantic formalism. However, prior knowledge of the English language can be encoded in these specified production rules, which dramatically improves the statistical efficiency of our model and obviates the need for massive training sets to learn English syntax. It is nonetheless tedious to design these rules while maintaining domain-generality. Once specified, however, these rules in principle can be re-used in new tasks and domains without further changes.
\end{enumerate}
We emphasize that only the prior is specified here, and our algorithm uses grammar induction to infer the posterior. In principle, a more relaxed choice of $H$ may enable grammar induction without pre-specified production rules, and therefore without dependence on a particular semantic formalism or natural language, if an efficient inference algorithm can be developed in such cases.

\subsection{Modeling morphology}

The grammar model is easily modified to incorporate word morphology. To do so, we add an additional step to the generative process after generating the terminal symbols. Instead of the terminal symbols constituting the tokens of the sentence directly, the terminal symbols are instead word roots coupled with morphological flags that indicate their inflection. For example, in the grammar in figure \ref{fig:example_grammar}, rather than having multiple rules for the various inflections of the verb ``to border'', such as $\textsf{V}\to\text{``borders''}$, $\textsf{V}\to\text{``bordered''}$, $\textsf{V}\to\text{``bordering''}$, there would only be a single production rule for the root: $\textsf{V}\to\text{``border''}$. In order to produce the various inflections, the logical form is augmented to carry morphology information.  Semantic transformation functions may add or modify morphological flags. For example, suppose we have the rule $\textsf{VP} \to \textsf{V\hspace{0.4mm}:\hspace{0.4mm}\texttt{\color{RoyalBlue}add\_third\_person,add\_present\_tense}}$ where \texttt{\color{RoyalBlue}add\_third\_person} is a function that adds the \textsc{\color{RedOrange!65!red!90!black}3rd} flag (indicating third person) to the logical form, and \texttt{\color{RoyalBlue}add\_present\_tense} is a function that adds the \textsc{\color{RedOrange!65!red!90!black}pres} flag (indicating present tense) to the logical form. These morphological flags are copied into the terminal symbols, and as a final step, the roots are inflected according to the morphological flags (e.g. ``border[\textsc{\color{RedOrange!65!red!90!black}3rd},\textsc{\color{RedOrange!65!red!90!black}pres}]'' is inflected to ``borders''). See figure \ref{fig:example_derivation_morph} for an example of a derivation tree for a grammar that models morphology.

\begin{figure}
	\definecolor{lightgrey}{rgb}{0.95,0.95,0.95}
	\centering
	\tikzstyle{nonterminal}=[circle,draw,fill=lightgrey,inner sep=0,minimum size=7mm,thick]
	\begin{tikzpicture}[
		level 1/.style={sibling distance=3.2cm,level distance=8mm,line width=1pt,edge from parent path={(\tikzparentnode) -- (\tikzchildnode)}},
		level 2/.style={sibling distance=2.3cm,level distance=8mm,line width=1pt,edge from parent path={(\tikzparentnode) -- (\tikzchildnode)}}]
	\node[nonterminal,label=right:\small{\color{RoyalBlue}\texttt{borders(pa,nj)}}] (S) {\textsf{\textbf{S}}}
		child {node[nonterminal,label=left:\small{\color{RoyalBlue}\texttt{pa}}] (N) {\textsf{\textbf{N}}}
	    	child {node[text depth=0mm] (pa) {\textrm{``Pennsylvania''}}}
		}
		child {node[nonterminal,label=right:\small{\color{RoyalBlue}\texttt{borders(,nj)}}] (VP) {\textsf{\textbf{VP}}}
			child {node[nonterminal] (V) {\textsf{\textbf{V}}}
				child {node[text height=2mm,text depth=0mm] (border) {\textrm{``border''[\textsc{\color{RedOrange!65!red!90!black}3rd},\textsc{\color{RedOrange!65!red!90!black}pres}]}}
				    child {node[text height=2mm,text depth=2mm] (borders) {\textrm{``borders''}}}
				}
			}
			child {node[nonterminal,label=right:\small{\color{RoyalBlue}\texttt{nj}}] (Nobj) {\textsf{\textbf{N}}}
				child {node[text height=2mm,text depth=0mm] (nj) {\textrm{``NJ''}}}
			}
		};
	\end{tikzpicture}
	\caption{Example of a derivation tree under a grammar with a model of morphology. The logical form corresponding to every node is shown in blue beside the respective node. The logical form for V is {\normalfont{\color{RoyalBlue}\texttt{borders(,nj)}}[\textsc{\color{RedOrange!65!red!90!black}3rd},\textsc{\color{RedOrange!65!red!90!black}pres}]} and is omitted to reduce clutter. Morphology is not modeled for proper nouns such as ``Pennsylvania'' and ``NJ.''}
	\label{fig:example_derivation_morph}
\end{figure}

If a root with morphological flags has multiple inflections, such as ``octopus''[\textsc{\color{RedOrange!65!red!90!black}pl}] (\textsc{\color{RedOrange!65!red!90!black}pl} indicates plural), the generative process selects one uniformly at random. During inference (i.e. parsing), this morphological model has the effect of performing morphological and syntactic-semantic parsing jointly, as we will show in the next section. Wiktionary \citep{Wiktionary} provides comprehensive high-quality morphology information for English verbs, common nouns, adjectives, and adverbs. Our implementation uses Wiktionary to construct a mapping between uninflected roots and inflected words, which is used in both directions: (1) given root and morphological flags, find the corresponding set of inflections, or (2) given an inflected word, find the corresponding set of roots and morphological flags. Note that only (2) is necessary for parsing and training, whereas (1) is necessary for generation.

Although this morphology model is implemented in our code, we do not use it in our experiments on \textsc{GeoQuery} and \textsc{Jobs}. The morphology model is used in the experiments described later in the thesis of \citet{SaparovThesis2022}.

\section{Inference and implementation} \label{sec:parser_inference}

\subsection{Training} \label{sec:induction}

\abucomment{Abu: Tom, i added the following paragraph to hopefully better introduce this section; how does it look?}

In this section, we describe how to infer the latent derivation trees $\bm{t} \triangleq \{t_1,\hdots,t_n\}$, given a collection of sentences $\bm{y} \triangleq \{y_1,\hdots,y_n\}$ and logical form labels $\bm{x} \triangleq \{x_1,\hdots,x_n\}$, where each derivation tree $t_i$ is distributed according to the conditional distribution described by the generative process in section \ref{sec:language_module_generative_process} above.

We describe grammar induction independently of the choice of rule distribution. We wish to compute the posterior $p(\bm{t} \mid \bm{x}, \bm{y})$ of the latent derivation trees. This is intractable to compute exactly, and so we resort to MCMC. To perform blocked Gibbs sampling, we pick initial values for $\bm{t}$ and repeat the following: For $i=1,\hdots,n$, sample $t_i \mid \bm{t}_{-i}, x_i, y_i$ where $\bm{t}_{-i} = \bm{t} \setminus \{ t_i \}$.
\begin{equation}
	p(t_i \mid \bm{t}_{-i}, x_i, y_i) \propto \mathds{1}\{\text{yield}(t_i) \hspace{-0.1em}=\hspace{-0.1em} y_i\} \hspace{-1mm} \prod_{A\in\mathcal{N}} \hspace{-1mm} p\Bigg( \hspace{-1mm} \bigcap_{\substack{\{\textbf{n} \in t_i \hspace{0.5mm} : \hspace{0.5mm} \textbf{n} \\ \text{\tiny has label } A\}}} \hspace{-2mm} r^{\textbf{n}} \hspace{0.3em}\Bigg|\hspace{0.3em} \bm{t}_{-i}, x_i \Bigg),
\end{equation}
where $\mathcal{N}$ is the set of nonterminals, and the intersection is taken over the nodes $\textbf{n} \in t_i$ labeled with the nonterminal $A$ in the $i^{\scriptsize th}$ derivation tree, and $r^{\textbf{n}}$ is the production rule at node $\textbf{n}$. Note that the probability does not necessarily factorize over rules, as is the case when using the HDP. So in order to sample $t_i$, we use Metropolis-Hastings (MH), where the proposal distribution is given by the fully factorized form:
\begin{align}
	p(t^*_i &\mid \bm{t}_{-i}, x_i, y_i) \propto \mathds{1}\{\text{yield}(t^*_i) \hspace{-0.1em}=\hspace{-0.1em} y_i\} \prod_{\textbf{n} \in t^*_i} p\left( r^{\textbf{n}} \mid \bm{t}_{-i}, x^{\textbf{n}}_i \right). \label{eq:factorized_joint}
\end{align}
The algorithm for sampling $t^*_i$ is detailed in section \ref{sec:sampling_derivation_trees}. After sampling $t^*_i$, we choose to accept the new sample with probability
\begin{equation}
	\min\left\{ 1, \frac{ \prod_{\textbf{n}\in t_i} p(r^{\textbf{n}} \mid x^{\textbf{n}},\bm{t}_{-i}) }{ p\left( \bigcap_{\textbf{n}\in t_i} r^{\textbf{n}} \mid x,\bm{t}_{-i} \right) } \frac{ p\left( \bigcap_{\textbf{n}\in t^*_i} r^{\textbf{n}} \mid x,\bm{t}_{-i} \right) }{ \prod_{\textbf{n}\in t^*_i} p(r^{\textbf{n}} \mid x^{\textbf{n}},\bm{t}_{-i}) } \right\}, \label{eq:derivation_tree_acceptance_probability}
\end{equation}
where $t_i$, here, is the old sample, and $t^*_i$ is the newly proposed sample. In practice, this acceptance probability is very high. This approach is very similar in structure to that in \citet{DBLP:conf/naacl/JohnsonGG07,DBLP:conf/naacl/BlunsomC10,DBLP:journals/jmlr/CohnBG10}.

Computing the conditional probabilities of the production rules $p(r^{\textbf{n}} \mid x^{\textbf{n}},\bm{t}_{-i})$ and $p( \bigcap_{\textbf{n}\in t_i} r^{\textbf{n}} \mid x,\bm{t}_{-i})$ (as well as the quantities required in sampling $t^*_i$) depends on the model for selecting production rules. In our semantic parsing model, which uses an HDP model, these quantities can be computed using equations \ref{eq:hdp_posterior_predictive_probability} and \ref{eq:hdp_joint_posterior_predictive_probability}. Our parsing method only keeps the last MCMC sample ($N_{\text{samples}}$ = 1), so for each node in every derivation tree $\textbf{m}\in t_j$, the production rule at that node $r^{\textbf{m}}$ corresponds to a customer in the Chinese restaurant representation of the HDP associated with the nonterminal at node $\textbf{m}$. When resampling the derivation tree $t_i$, our method removes all customers that correspond to a production rule in $t_i$. Then it is straightforward to compute conditional probabilities according to equations \ref{eq:hdp_posterior_predictive_probability} and \ref{eq:hdp_joint_posterior_predictive_probability} with the remaining customers. Once a new $t_i$ is sampled, the customers corresponding to production rules in $t_i$ are added to their respective restaurants.

There may be additional random variables in the grammar apart from the derivation trees, such as $\bm{\alpha}$ in the HDPs. We perform Gibbs sampling steps for these variables after each loop of resampling the trees $t_i$, $i=1,\hdots,n$. The grammar induction algorithm is summarized: Pick initial values for $\bm{t}$ and $\bm{\alpha}$ and repeat the following,
\begin{enumerate}
    \item For $i=1,\hdots,n$, sample $t_i^* \mid \bm{\alpha},\bm{t}_{-i},x_i,y_i$ from the distribution given by equation \ref{eq:factorized_joint}. Then accept this sample as the new value for $t_i$ with probability given by equation \ref{eq:derivation_tree_acceptance_probability}.
    \item Perform the Gibbs sampling step for $\bm{\alpha} \mid \bm{t}$.
\end{enumerate}
In all our experiments, we run the above loop for 10 iterations. Note that this algorithm requires no further supervision beyond the utterances $\bm{y}$ and logical forms $\bm{x}$. However, it is able to exploit additional information such as supervised derivation trees: if $\bar{\bm{t}} \subseteq \bm{t}$ is a subset of derivation trees that are supervised, the Gibbs sampling algorithm simply avoids resampling the trees in $\bar{\bm{t}}$. These supervised derivation trees do not necessarily need to be rooted in the nonterminal \textsf{S}. For example, a lexicon can be provided where each entry is a terminal symbol $y_i$ with a corresponding logical form label $x_i$. In our experiments on \textsc{GeoQuery} and \textsc{Jobs}, we evaluate our method with and without such a lexicon.

\subsubsection{Sampling $t^*_i$} \label{sec:sampling_derivation_trees}

To sample from equation \ref{eq:factorized_joint}, we use \emph{inside-outside sampling} \citep{DBLP:conf/emnlp/FinkelMN06,DBLP:conf/naacl/JohnsonGG07}, a dynamic programming approach. For every nonterminal $A\in\mathcal{N}$, sentence start position $i$, end position $j$, and logical form $x$, let $I_{(A, i, j, x)}$ be the probability that $t^*_i$ has a node $\textbf{n}$ with the label $A$ and logical form $x$ and spans the sentence from $i$ to $j$. This is known as the \emph{inside probability}. Similarly, for all production rules in the grammar $A \to B_1\hspace{-1mm}\nobreak:\nobreak\hspace{-1mm}f_1 \hdots B_K\hspace{-1mm}\nobreak:\nobreak\hspace{-1mm}f_K$, sentence boundary positions between the right-hand side nonterminals $l_1<\hdots<l_{K+1}$, and logical forms $x$, let $I_{(\textsf{S}\to B_1\nobreak:\nobreak f_1 \hdots B_K\nobreak:\nobreak f_K,\bm{l},x)}$ be the probability that $t^*_i$ has a node $\textbf{n}$ with the label $A$ and logical form $x$ and has child nodes $B_u$, each with logical forms $f_u(x)$, and each spanning the sentence from $l_u$ to $l_{u+1}$. This is known as the \emph{inside rule probability}.
Note that we don't need to compute all possible inside probabilities for all logical forms (in many applications, the set of logical forms is infinite). Therefore, we compute these inside probabilities top-down, beginning at the root nonterminal $I_{(\textsf{S},0,|y_i|,x_i)}$ with the known logical form $x_i$ where $|y_i|$ is the length of sentence $y_i$. The following formula can be used to compute this quantity recursively:
\begin{align}
	I_{(A,i,j,x)} &= \hspace{-1.6em}\sum_{A\to B_1:f_1 \hdots B_K:f_K} \hspace{0.4em}\sum_{i=l_1<\hdots<l_{K+1}=j} \hspace{-1.6em} I_{(A\to B_1\nobreak:\nobreak f_1 \hdots B_K\nobreak:\nobreak f_K,\bm{l},x)}. \\
    I_{(A\to B_1\nobreak:\nobreak f_1 \hdots B_K\nobreak:\nobreak f_K,\bm{l},x)} &= p(A\to B_1\hspace{-1mm}\nobreak:\nobreak\hspace{-1mm}f_1 \hdots B_K\hspace{-1mm}\nobreak:\nobreak\hspace{-1mm}f_K \mid x,\bm{t}_{-i}) \prod_{u=1}^K \hspace{1mm} I_{(B_u,l_u,l_{u+1},f_u(x))}.
\end{align}
If $f_u(x)$ returns failure, then $I_{(B_u,l_u,l_{u+1},f_u(x))}=0$. Note that in the case that $A$ is a preterminal,
\begin{equation}
    \hspace{-0.3em}I_{(A\to w,l_1,l_2,x)} = \mathds{1}\{w \text{ matches } y_i \text{ at } (l_1,l_2)\} \hspace{0.25em} p(A\to w \mid \bm{t}_{-i}).
\end{equation}
where $w$ is a terminal. Aside from the inside probabilities that were required to compute the root inside probability $I_{(\textsf{S},0,|y_i|,x_i)}$, all other inside probabilities are $0$. In our code, this recursion is implemented iteratively, in order to avoid any issues with limited stack size and to share code with the parsing and generation algorithms. We also take care not to recompute previously computed inside probabilities.

All that remains is the outside step: sample the derivation tree using the computed inside probabilities. To do so, start with the root nonterminal \textsf{S} at positions $i=0$ to $j=|y_i|$ and logical form $x_i$, and consider all production rules with \textsf{S} on the left-hand side $\textsf{S}\to B_1\hspace{-1mm}\nobreak:\nobreak\hspace{-1mm}f_1 \hdots B_K\hspace{-1mm}\nobreak:\nobreak\hspace{-1mm}f_K$ and all sentence boundaries between the right-hand side nonterminals $l_1<\hdots<l_K<l_{K+1}$ where $l_1=i$ and $l_{K+1}=j$. Sample a production rule and sentence boundaries with probability proportional to the inside rule probability $\smash{I_{(\textsf{S}\to B_1\nobreak:\nobreak f_1 \hdots B_K\nobreak:\nobreak f_K,\bm{l},x_i)}}$. Next, consider each right-hand side nonterminal of the selected rule $B_u$, start position $l_u$, end position $l_{u+1}$, and logical form $f_u(x_i)$, and recursively repeat the sampling procedure. The end result is a tree sampled from equation \ref{eq:factorized_joint}.

\subsection{Parsing} \label{sec:parsing}

For a new sentence $y_*$, we aim to find the logical form $x_*$ and derivation $t_*$ that maximizes
\begin{align}
    p(x_*,t_* \mid y_*,\bm{x},\bm{y}) &= \int p(x_*,t_* \mid y_*,\bm{t}) p(\bm{t} \mid \bm{x},\bm{y}) d\bm{t}, \\
        &\approx \frac{1}{N_\text{samples}} \sum_{\bm{y}\sim \bm{t} \mid \bm{x},\bm{y}} p(x_*,t_* \mid y_*,\bm{t}).
\end{align}
These samples of $\bm{t}$ are obtained from the above training procedure. For the parsing approach presented in this section, it is assumed that $N_\text{samples}=1$, and so $p(x_*,t_* \mid y_*,\bm{x},\bm{y}) \approx p(x_*,t_* \mid y_*,\bm{t})$, where $\bm{t}$ is the last MH sample from the training procedure. Thus, we can write the objective function for parsing:
\begin{align}
	p(x_*,t_* \mid y_*,\bm{t}) &\propto p(x_*) p(y_* \mid t_*) p(t_* \mid x_*,\bm{t}), \nonumber \\
		&= \mathds{1}\{\text{yield}(t_*)=y_*\} p(x_*) p\bigg( \bigcap_{\textbf{n} \in t_*} r^{\textbf{n}} \hspace{0.3em}\bigg|\hspace{0.3em} x^{\textbf{n}}_*, \bm{t} \bigg), \label{eq:exact_objective} \\
		&\approx \mathds{1}\{\text{yield}(t_*)=y_*\} p(x_*) \prod_{\textbf{n} \in t_*} p(r^{\textbf{n}} \mid x^{\textbf{n}}_*, \bm{t}).\footnotemark \label{eq:objective}
\end{align}
This\footnotetext{Equations \ref{eq:exact_objective} and \ref{eq:objective} are not equal since the conditional distribution of production rules is not necessarily i.i.d. Our semantic parsing model uses an HDP for this conditional distribution, which has the nice property that as the size of the training set increases, this approximation becomes more exact.} is a discrete optimization problem, which we solve using \emph{branch-and-bound} (see algorithm \ref{alg:branch_and_bound}). The algorithm starts by considering the set of all derivation trees of $y_*$ and partitions it into a number of subsets (the ``branch'' step). For each subset $S$, we compute an upper bound on the log probability of any derivation in $S$ (the ``bound'' step). This bound is given by equations \ref{eq:priority}, \ref{eq:priority2}, and \ref{eq:priority3}. Having the computed the bound for each subset, we push them onto a priority queue, prioritized by the bound. We then pop the subset with the highest bound and repeat this process, further subdividing this set into subsets, computing the bound for each subset, and pushing them onto the queue. Eventually, we will pop a subset containing a single derivation which is provably optimal, if its objective function value according to the above equation is at least the priority of the next item in the queue. We can continue the algorithm to obtain the top-$k$ derivations/logical forms. Since this algorithm operates over \emph{sets} of logical forms (where each set is possibly infinite), we must implement a data structure to sparsely represent such sets of formulas, as well as algorithms to perform set operations, such as intersection and subtraction.

Each set of derivations is sparsely represented in our implementation as a single incomplete derivation tree (i.e. the leaf nodes may be either terminals or nonterminals) and a logical form set. The logical form set represents the logical form of the root node of every derivation tree in the set. The logical forms at the other nodes can be computed by using the semantic transformation functions. In addition, every nonterminal node with non-zero children has two integer indices that indicate its start and end positions in the sentence $y_*$. This data structure represents the set of all derivation trees whose nodes match the nodes in the incomplete derivation tree at the given sentence positions. In addition, each derivation tree set has an integer counter to indicate to the \texttt{branch} function how to subdivide the set. Each such set of derivation trees is also called a \emph{search state}. As an example, consider the input sentence ``Trenton is the capital of New Jersey.'' Now consider a search state where the incomplete derivation tree contains only a single node labeled \textsf{NP} with start position 3 and end position 7 (corresponding to ``capital of New Jersey'') and the logical form set is the set of all logical forms. This search state represents the set of all derivation trees that have a node with label \textsf{NP} and is the common ancestor of the terminals in ``capital of New Jersey.''

Given a set of derivation trees, the \texttt{branch} function is defined in algorithm \ref{alg:parser_branch}. The branch-and-bound algorithm is started with a derivation tree set whose incomplete derivation tree has a single root node with nonterminal \textsf{S}, the set of all logical forms, start position $0$, and end position $|y_*|$.

{\setlength{\algomargin}{0em}
\begin{algorithm2e}
\footnotesize
\let\oldnl\nl
\newcommand{\nonl}{\renewcommand{\nl}{\let\nl\oldnl}}
\SetNlSty{}{\color{RedOrange}\sffamily}{}
\SetAlgoBlockMarkers{}{}
\SetKwProg{Fn}{function}{}{}
\SetKwIF{If}{ElseIf}{Else}{if}{ }{else if}{else }{}
\SetKw{Continue}{continue}
\SetKwFunction{FBranch}{\small branch}
\SetKwFunction{FExpand}{\small expand}
\SetKwFunction{FComplete}{\small complete}
\SetKwFor{ForEach}{for each}{do}{end}
\SetKwProg{uForEach}{for each}{ do}{}
\SetKwProg{Fn}{function}{}{}
\AlgoDisplayBlockMarkers\SetAlgoVlined
\SetAlCapNameFnt{\small}
\SetAlCapFnt{\small}
\SetNoFillComment
\DontPrintSemicolon
\SetInd{0.0em}{0.8em}
    \Fn{\FBranch{derivation tree set $S$}}{
        $L$ is an empty list \;
        $\textbf{n}$ is the root of the incomplete derivation tree of $S$ \;
        $X$ is the logical form set at $\textbf{n}$ \;
        $i$ is the start sentence position of $\textbf{n}$ \;
        $j$ is the end sentence position of $\textbf{n}$ \;
        \uIf{\textbf{n} has no child nodes}{
            $A$ is the nonterminal symbol of $\textbf{n}$ \;
            \Return{$\texttt{expand(} A, i, j, X \texttt{)}$} \tcc{see algorithm \ref{alg:parser_expand}}
        }\uElseIf{$\textbf{n}$ has a nonterminal child node with no children}{
            $A\to B_1\hspace{-1mm}\nobreak:\nobreak\hspace{-1mm}f_1 \hdots B_K\hspace{-1mm}\nobreak:\nobreak\hspace{-1mm}f_K$ is the production rule at $\textbf{n}$ \;
            $\textbf{c}_k$ is the first nonterminal child node of $\textbf{n}$ with no children \;
            $B_k$ is the nonterminal symbol of $\textbf{c}_k$ \;
            $i_k$ is the start sentence position of $\textbf{c}_k$ \;
            $j_k$ is the end sentence position of $\textbf{c}_k$ \;
            $m$ is the counter of $S$ \;
            $S_m$ is the $m^{\scriptsize th}$ most probable set of derivation trees with root nonterminal $B_k$, start position $i_k$, end position $j_k$, whose logical forms are a subset of $f_k(X) \triangleq \{f_k(x) \ne \textit{fail} : x\in X\}$, according to equation \ref{eq:objective} \label{line:parser_branch_recursion_step} \;
            \If{$S_m$ exists}{
                \For{sentence positions $j_{k+1}$ such that $j_k < j_{k+1} < j$}{
                    \tcc{the operation $X \cap f_k^{-1}(X_m)$ can return a union of sets}
                    let $X_{k,1} \cup \hdots \cup X_{k,r}$ be the output of $X \cap f_k^{-1}(X_m)$ where $X_m$ is the logical form set of $S_m$, and $f_k^{-1}(X_m) \triangleq \{x : f_k(x) \in X_m\}$ \label{line:parser_invert_intersect} \;
                    \For{$X_{k,l}\in\{X_{k,1},\hdots,X_{k,r}\}$}{
                        $S^*$ is a new derivation tree set with counter $1$, the incomplete derivation tree is identical to that of $S$ except $\textbf{c}_k$ is substituted with the incomplete derivation tree of $S_m$, the logical form set at the root is $X_{k,l}$, and the end position of $\textbf{c}_{k+1}$ is $j_{k+1}$ \label{line:parser_branch_substitute_child_subtree} \;
                        $L\texttt{.add(} S^* \texttt{)}$ \;
                    }
                }
                $L\texttt{.add(}$ a new derivation tree set identical to $S$ except its counter is $m+1 \texttt{)}$ \;
            }
        }\Else{
            $A\to B_1\hspace{-1mm}\nobreak:\nobreak\hspace{-1mm}f_1 \hdots B_K\hspace{-1mm}\nobreak:\nobreak\hspace{-1mm}f_K$ is the production rule at $\textbf{n}$ \;
            $m$ is the counter of $S$ \;
            $X_m$ is the $m^{\scriptsize th}$ most likely set of logical forms according to $p(A\to B_1\hspace{-1mm}\nobreak:\nobreak\hspace{-1mm}f_1 \hdots B_K\hspace{-1mm}\nobreak:\nobreak\hspace{-1mm}f_K \mid x\in X, \bm{t})$ \label{line:parser_branch_complete_step} \;
            \If{$X_m$ exists}{
                $L\texttt{.add(}$ a new derivation tree set identical to $S$ except its logical form set is $X_m$, and is marked as \texttt{COMPLETE} $\texttt{)}$ \label{line:parser_branch_complete_derivation} \;
                $L\texttt{.add(}$ a new derivation tree set identical to $S$ except its counter is $m+1 \texttt{)}$ \;
            }
        }
        \Return{$L$}
    }
    \caption{Pseudocode for \texttt{branch} in the branch-and-bound algorithm for the parser, which aims to maximize equation \ref{eq:objective}.}
    \label{alg:parser_branch}
\end{algorithm2e}}

{\setlength{\algomargin}{0em}
\begin{algorithm2e}
\footnotesize
\let\oldnl\nl
\newcommand{\nonl}{\renewcommand{\nl}{\let\nl\oldnl}}
\SetNlSty{}{\color{RedOrange}\sffamily}{}
\SetAlgoBlockMarkers{}{}
\SetKwProg{Fn}{function}{}{}
\SetKwIF{If}{ElseIf}{Else}{if}{ }{else if}{else }{}
\SetKw{Continue}{continue}
\SetKwFunction{FBranch}{\small branch}
\SetKwFunction{FExpand}{\small expand}
\SetKwFunction{FComplete}{\small complete}
\SetKwFor{ForEach}{for each}{do}{end}
\SetKwProg{uForEach}{for each}{ do}{}
\SetKwProg{Fn}{function}{}{}
\AlgoDisplayBlockMarkers\SetAlgoVlined
\SetAlCapNameFnt{\small}
\SetAlCapFnt{\small}
\SetNoFillComment
\DontPrintSemicolon
\SetInd{0.0em}{0.8em}
    \Fn{\FExpand{nonterminal $A$, start position $i$, end position $j$, logical form set $X$}}{
        $L$ is an empty list \;
        \uIf{$A$ is a preterminal}{
            \For{rules $A\to w$ where the tokens in the sentence $y_*$ matches the terminal $w$ at positions $i$ to $j$}{
                \tcc{if using the morphology model, we instead require that $w$ is a valid morphological parse of the tokens in the sentence $y_*$ at positions $i$ to $j$}
                $S^*$ is a new derivation tree set where the incomplete derivation tree consists of a root node \textbf{n} with nonterminal $A$, start position $i$, end position $j$, logical form set $X$, and child node $w$ \;
                $L\texttt{.add(} S^* \texttt{)}$ \;
            }
        }\Else{
            \For{rules $A\to B_1\hspace{-1mm}\nobreak:\nobreak\hspace{-1mm}f_1 \hdots B_K\hspace{-1mm}\nobreak:\nobreak\hspace{-1mm}f_K$ and sentence positions $k$ such that $i<k<j$}{
                $S^*$ is a new derivation tree set with counter $1$, the incomplete derivation tree consists of a root node $\textbf{n}$ with nonterminal $A$, start position $i$, end position $j$, logical form set $X$, and for each child node $\textbf{c}_i$, the nonterminal is $B_i$, and logical form set is $f_i(X) \triangleq \{f_i(x) \ne \textit{fail} : x\in X\}$; the start position of $\textbf{c}_1$ is $i$, the end position of $\textbf{c}_1$ is $k$, and the end position of $\textbf{c}_K$ is $j$ \;
                $L\texttt{.add(} S^* \texttt{)}$ \;
            }
        }
        \Return{$L$}
    }
    \caption{Pseudocode for the \texttt{expand} helper function, which algorithm \ref{alg:parser_branch} invokes.}
    \label{alg:parser_expand}
\end{algorithm2e}}

{\setlength{\algomargin}{0em}
\begin{algorithm2e}
\footnotesize
\let\oldnl\nl
\newcommand{\nonl}{\renewcommand{\nl}{\let\nl\oldnl}}
\SetNlSty{}{\color{RedOrange}\sffamily}{}
\SetAlgoBlockMarkers{}{}
\SetKwProg{Fn}{function}{}{}
\SetKwIF{If}{ElseIf}{Else}{if}{ }{else if}{else }{}
\SetKw{Continue}{continue}
\SetKwFunction{FGetKBest}{\small get\_kth\_best}
\SetKwFor{ForEach}{for each}{do}{end}
\SetKwProg{uForEach}{for each}{ do}{}
\SetKwProg{Fn}{function}{}{}
\AlgoDisplayBlockMarkers\SetAlgoVlined
\SetAlCapNameFnt{\small}
\SetAlCapFnt{\small}
\SetNoFillComment
\DontPrintSemicolon
\SetInd{0.0em}{0.8em}
    \Fn{\FGetKBest{objective function $f$, \newline \phantom{\hspace{11.1em}} heuristic $h$, \newline \phantom{\hspace{11.1em}} priority queue $Q$, \newline \phantom{\hspace{11.1em}} list of completed elements $C$, \newline \phantom{\hspace{11.1em}} integer $k$}}{
        \If{there are at least $k$ elements in $C$ with priority at least the highest priority in $Q$}{
            \Return{$k$-best element in $C$}
        }
        \While{$Q$ not empty}{
            $(S,v) = Q\texttt{.pop}()$ \;
            \uIf{$S$ is marked as \texttt{COMPLETE}}{
                $C\texttt{.add}(x, f(x))$ \;
            }
            \Else{
                $(S_1,\hdots,S_n) = \texttt{branch}(S)$ \;
                \For{$i = 1,\hdots,n$}{
                    $Q\texttt{.push}(S_i, h(S_i))$ \;
                }
            }

            \tcc{check termination condition}
            \If{there are at least $k$ elements in $C$ with priority at least $v$}{
                \Return{$k$-best element in $C$}
            }
        }
        \Return{$\varnothing$}
    }
    \caption{A modified branch-and-bound algorithm to return the $k^{\scriptsize th}$ best element(s) that maximize(s) the function $f$. Before the first call to this function, $C$ is initialized as an empty list, and $Q$ is initialized with a single element: $Q\texttt{.push(} X, h(X) \texttt{)}$ where $X$ is the domain on which to maximize $f$. The changes to $C$ and $Q$ persist across subsequent calls to $\texttt{get\_kth\_best}$.}
    \label{alg:get_kth_best}
\end{algorithm2e}}

There are two missing pieces in algorithm \ref{alg:parser_branch}: the first is on line \ref{line:parser_branch_recursion_step}. To compute this, we can augment the branch-and-bound algorithm to return the $m^{\scriptsize th}$ best element(s) that maximize(s) an objective function over a set. This augmented function is shown in algorithm \ref{alg:get_kth_best}. Whenever line \ref{line:parser_branch_recursion_step} is first executed for a given nonterminal $B_k$, start position $i_k$, end position $j_k$ and logical form set $f_k(X)$, initialize the priority queue $Q$ in algorithm \ref{alg:get_kth_best} with: $Q\texttt{.push(} S^*, h(S^*) \texttt{)}$ where $S^*$ is the search state with an incomplete derivation tree consisting of a single node at the root with nonterminal $B_k$, start position $i_k$, end position $j_k$, and logical form set $f_k(X)$.

The other missing piece in algorithm \ref{alg:parser_branch} is line \ref{line:parser_branch_complete_step}, which depends on the model for selecting production rules. Our semantic parsing model uses an HDP model, and is able to directly use the algorithm described in section \ref{sec:inferring_hdp_source_node} to compute $X^*$. Algorithm \ref{alg:get_kth_best} may also be used here to return the $m^{\scriptsize th}$ most likely logical form(s).

The above branch-and-bound algorithm requires a heuristic function that, for an input search state (set of derivation trees), returns an upper bound on the objective function in equation \ref{eq:objective} over all derivation trees in the set. This heuristic function determines the order of the search states to visit. The product in the objective $\prod_{\textbf{n} \in t_*} p(r^{\textbf{n}} \mid x^{\textbf{n}}_*, \bm{t})$ can be decomposed accordingly into a product of two components: (1) the \emph{inner probability} at a node $\textbf{n}\in t^*$ is the product of the terms that correspond to the subtree rooted at $\textbf{n}$, and (2) the \emph{outer probability} is the product of the remaining terms, which correspond to the parts of $t_*$ outside of the subtree rooted at $\textbf{n}$.

To help define this heuristic, we define an upper bound on the log inner probability $I_{(A,i,j)}$ for any derivation tree rooted at nonterminal $A$ at start position $i$ and end position $j$ in the sentence.
\begin{align}
	&\hspace{-2.2em}I_{(A,i,j)} \triangleq \max_{A\to B_1\hdots B_K} \bigg( \nonumber \\
		&\hspace{3.0em}\max_{x'} \log p(A \hspace{-0.2em}\to\hspace{-0.2em} B_1,\hdots,B_K \mid x',\bm{t}) + \max_{l_2<\hdots<l_K} \sum_{k=1}^K I_{(B_k,l_k,l_{k+1})} \bigg),
\end{align}
where $l_1 = i$, $l_{K+1} = j$. Note that the left term is a maximum over all logical forms $x'$, and so this upper bound only considers syntactic information. Computing the left term depends on the model for selecting production rules. Since our semantic parsing model uses an HDP, it uses the branch-and-bound approach in section \ref{sec:inferring_hdp_source_node} to compute this term. The right term can be maximized using dynamic programming with running time $\mathcal{O}(K^2)$. As such, classical syntactic parsing algorithms can be applied to compute $I$ for every chart cell in $\mathcal{O}(n^3)$. For any terminal symbol $w\in\mathcal{W}$, we define $I_{(w,i,j)} = 0$.

\begin{figure}
\definecolor{sand}{rgb}{0.94,0.80,0.67}
\definecolor{darkergray}{rgb}{0.6,0.6,0.6}
\definecolor{gray}{rgb}{0.84,0.84,0.84}
\hspace*{0.5\textwidth} \makebox[\paperwidth]{%
\begin{tikzpicture}[level distance=4.7mm]
	\scriptsize

	\node[node distance=0mm,rounded corners=2pt,fill=sand,align=center,text width=31.7mm,inner sep=1mm,sibling distance=7mm] (root_state) {
		\tikzset{edge from parent/.style={draw,edge from parent path={(\tikzparentnode.south) -- +(0,-0.8mm) -| (\tikzchildnode)}},rounded corners=0pt,text width={},inner sep=0.4mm}
		\hspace{-0.85em}\texttt{\color{RoyalBlue}*}
		\hspace{0.3em} \Tree [.\textsf{S} [.\text{*} \text{``Pennsylvania borders NJ''} ] ] \raisebox{-1.5em}{\hspace{-0.15em}\Bigg\}} \\[0.1em]
		upper bound: \textbf{-5.26}
	};
	\node[right=8em of root_state.north,yshift=-0.56em] (logical_form_label) {\scriptsize\color{black!65} set of logical forms in this search state};
	\draw [-,color=black!65] (logical_form_label) to ([xshift=0.2em,yshift=-0.56em]root_state.north);
	\node[right=8em of root_state.north,yshift=-3.34em] (derivation_label) {\scriptsize\color{black!65} set of derivation trees};
	\draw [-,color=black!65] (derivation_label) to ([xshift=6.72em,yshift=-3.34em]root_state.north);
	\node[right=8em of root_state.north,yshift=-6.95em,text width=16em] (bound_label) {\scriptsize\color{black!65} upper bound on log posterior of any derivation in this search state};
	\draw [-,color=black!65] (bound_label) to ([xshift=4.5em,yshift=-6.95em]root_state.north);

	\node[below left=2em of root_state,xshift=-0.5em,node distance=0mm,rounded corners=2pt,fill=gray,align=center,text width=14.1em,inner sep=1mm,sibling distance=0mm] (state01) {
		\tikzset{edge from parent/.style={draw,edge from parent path={(\tikzparentnode.south) -- +(0,-0.8mm) -| (\tikzchildnode)}},rounded corners=0pt,text width={},inner sep=0.4mm}
		\texttt{\color{RoyalBlue}*(*,*)} \\[1mm]
		\Tree [.\textsf{S} \edge;
				[.\textsf{N} [.\text{*} \text{``Pennsylvania borders''} ] ]
				[.\textsf{VP} [.\text{*} \text{``NJ''} ] ]
			] \\
		upper bound: \textbf{-12.98}
	};
	\node[right=1em of state01,node distance=0mm,rounded corners=2pt,fill=sand,align=center,text width=14.1em,inner sep=1mm,sibling distance=0mm] (state02) {
		\tikzset{edge from parent/.style={draw,edge from parent path={(\tikzparentnode.south) -- +(0,-0.8mm) -| (\tikzchildnode)}},rounded corners=0pt,text width={},inner sep=0.4mm}
		\texttt{\color{RoyalBlue}*(*,*)} \\[1mm]
		\Tree [.\textsf{S} \edge;
				[.\textsf{N} [.\text{*} \text{``Pennsylvania''} ] ]
				[.\textsf{VP} [.\text{*} \text{``borders NJ''} ] ]
			] \\
		upper bound: \textbf{-5.26}
	};
	\node[left=0.8em of state01,text width=27.2mm,yshift=5.65em] (branch0) {\scriptsize\color{black!65} branch according to production rules with \textsf{S} as left-hand side};
	\draw [-,dashed,darkergray] ([yshift=0.4em]branch0.east) to ([xshift=50.0em,yshift=0.4em]branch0.east);

	\draw [-{Latex[length=1.8mm,width=1.8mm]},thick] (root_state) to (state01);
	\draw [-{Latex[length=1.8mm,width=1.8mm]},thick] (root_state) to (state02);

	\node[below=2em of state01,node distance=0mm,rounded corners=2pt,fill=gray,align=center,text width=14.1em,inner sep=1mm,sibling distance=0mm] (state11) {
		\tikzset{edge from parent/.style={draw,edge from parent path={(\tikzparentnode.south) -- +(0,-0.8mm) -| (\tikzchildnode)}},rounded corners=0pt,text width={},inner sep=0.4mm}
		\texttt{\color{RoyalBlue}*(<entity of new type>,*)} \\[1mm]
		\Tree [.\textsf{S} \edge;
				[.\textsf{N} \text{``Pennsylvania borders''} ]
				[.\textsf{VP} [.\text{*} \text{``NJ''} ] ]
			] \\
		upper bound: \textbf{-12.98}
	};
	\node[right=1em of state11,node distance=0mm,rounded corners=2pt,fill=sand,align=center,text width=14.1em,inner sep=1mm,sibling distance=0mm] (state12) {
		\tikzset{edge from parent/.style={draw,edge from parent path={(\tikzparentnode.south) -- +(0,-0.8mm) -| (\tikzchildnode)}},rounded corners=0pt,text width={},inner sep=0.4mm}
		\texttt{\color{RoyalBlue}*(pa,*)} \\[1mm]
		\Tree [.\textsf{S} \edge;
				[.\textsf{N} \text{``Pennsylvania''} ]
				[.\textsf{VP} [.\text{*} \text{``borders NJ''} ] ]
			] \\
		upper bound: \textbf{-5.82}
	};
	\node[right=0.5em of state12,align=center] (dots1) {$\hdots$};
	\node[right=0.5em of dots1,node distance=0mm,rounded corners=2pt,fill=gray,align=center,text width=13.3em,inner sep=1mm,sibling distance=0mm] (state13) {
		\tikzset{edge from parent/.style={draw,edge from parent path={(\tikzparentnode.south) -- +(0,-0.8mm) -| (\tikzchildnode)}},rounded corners=0pt,text width={},inner sep=0.4mm}
		\texttt{\color{RoyalBlue}*(<new state>,*)} \\[1mm]
		\Tree [.\textsf{S} \edge;
				[.\textsf{N} \text{``Pennsylvania''} ]
				[.\textsf{VP} [.\text{*} \text{``borders NJ''} ] ]
			] \\
		upper bound: \textbf{-13.12}
	};
	\node[left=0.8em of state11,text width=27.2mm,yshift=5.8em] (branch1) {\scriptsize\color{black!65} branch according to derivation trees of first child (i.e. \textsf{N}, computed recursively)};
	\draw [-,dashed,darkergray] ([yshift=0.5em]branch1.east) to ([xshift=50.0em,yshift=0.5em]branch1.east);

	\draw [-{Latex[length=1.8mm,width=1.8mm]},thick] (state01) to (state11);
	\draw [-{Latex[length=1.8mm,width=1.8mm]},thick] (state02) to (state12);
	\draw [-{Latex[length=1.8mm,width=1.8mm]},thick] (state02) to (state13);

	\node[below=2em of state11,node distance=0mm,rounded corners=2pt,fill=sand,align=center,text width=14.1em,inner sep=1mm,sibling distance=0mm] (state21) {
		\tikzset{edge from parent/.style={draw,edge from parent path={(\tikzparentnode.south) -- +(0,-0.8mm) -| (\tikzchildnode)}},rounded corners=0pt,text width={},inner sep=0.4mm}
		\texttt{\color{RoyalBlue}borders(pa,nj)} \\[1mm]
		\Tree [.\textsf{S} \edge;
				[.\textsf{N} \text{``Pennsylvania''} ]
				[.\textsf{VP}
					[.\textsf{V} \text{``borders''} ]
					[.\textsf{N} \text{``NJ''} ]
				]
			] \\[1mm]
		upper bound: \textbf{-6.74}
	};
	\node[right=1em of state21,node distance=0mm,rounded corners=2pt,fill=gray,align=center,text width=14.1em,inner sep=1mm,sibling distance=0mm] (state22) {
		\tikzset{edge from parent/.style={draw,edge from parent path={(\tikzparentnode.south) -- +(0,-0.8mm) -| (\tikzchildnode)}},rounded corners=0pt,text width={},inner sep=0.4mm}
		\texttt{\color{RoyalBlue}borders(pa,red)} \\[1mm]
		\Tree [.\textsf{S} \edge;
				[.\textsf{N} \text{``Pennsylvania''} ]
				[.\textsf{VP}
					[.\textsf{V} \text{``borders''} ]
					[.\textsf{ADJ} \text{``NJ''} ]
				]
			] \\[1mm]
		upper bound: \textbf{-18.62}
	};
	\node[right=0.5em of state22,align=center] (dots2) {$\hdots$};
	\node[right=0.5em of dots2,node distance=0mm,rounded corners=2pt,fill=gray,align=center,text width=13.3em,inner sep=1mm,sibling distance=0mm] (state23) {
		\tikzset{edge from parent/.style={draw,edge from parent path={(\tikzparentnode.south) -- +(0,-0.8mm) -| (\tikzchildnode)}},rounded corners=0pt,text width={},inner sep=0.4mm}
		\texttt{\color{RoyalBlue}<new relation>(pa)} \\[1mm]
		\Tree [.\textsf{S} \edge;
				[.\textsf{N} \text{``Pennsylvania''} ]
				[.\textsf{VP}
					[.\textsf{V} \text{``borders NJ''} ]
				]
			] \\[1mm]
		upper bound: \textbf{-10.13}
	};
	\node[left=0.8em of state21,text width=27.2mm,yshift=6.45em] (branch2) {\scriptsize\color{black!65} branch according to derivation trees of second child (i.e. \textsf{VP}, computed recursively)};
	\draw [-,dashed,darkergray] ([yshift=0em]branch2.east) to ([xshift=50.0em,yshift=0em]branch2.east);

	\draw [-{Latex[length=1.8mm,width=1.8mm]},thick] (state12) to (state21);
	\draw [-{Latex[length=1.8mm,width=1.8mm]},thick] (state12) to (state22);
	\draw [-{Latex[length=1.8mm,width=1.8mm]},thick] (state12) to (state23);

\end{tikzpicture}
}\hspace*{-\paperwidth} \\
\caption{The search tree of the branch-and-bound algorithm during parsing. In this diagram, each block is a search state, which represents a set of derivation trees. The blue asterisk \texttt{\color{RoyalBlue}*} denotes the set of all possible logical forms, whereas the black asterisk * denotes the set of all possible derivation (sub)trees. Note only the logical form at the root node is shown. The gray-colored search states are unvisited by the parser, since their upper bounds on the log posterior are smaller than that of the completed parse at the bottom of the diagram (-6.74), thus allowing the parser to ignore a very large number of improbable logical forms and derivations. In this example, we use the grammar from figure \ref{fig:example_grammar}. The branching steps here are simplified for the sake of illustration. The recursive optimization of the derivation subtrees for \textsf{N} and \textsf{VP} are not shown, which have their own respective search trees. \tomcomment{Tom: this is the most useful figure in the chapter.  I believe it would work better to introduce this earlier, since it shows an actual parse. If you did that, you could ground many of your earlier discussions about probabilistic inference and sampling in terms of this figure and its example parse.  You probably don’t have time to do this before you need to send this to your committee, but consider doing it afterwards.} \abucomment{Abu: hmm, i copied it into chapter 2 as an attempt to try to dispel the notion that we use the generative process itself for parsing; do you think this is good?}} \label{fig:parsing_search_tree}
\end{figure}

We now define the upper bound heuristic on any search state $S$ with an incomplete derivation tree that has root node $\textbf{n}$, start position $i$, end position $j$, and logical form set $X$. If $\textbf{n}$ has no child nodes:
\begin{equation}
    \log h(S) \triangleq h_x^{\textbf{n}}(X) + I_{(A,i,j)}. \label{eq:priority}
\end{equation}
Else, if $\textbf{n}$ has a nonterminal child node without children, the production rule at $\textbf{n}$ is $A \to B_1\hspace{-1mm}\nobreak:\nobreak\hspace{-1mm}f_1 \hdots B_K\hspace{-1mm}\nobreak:\nobreak\hspace{-1mm}f_K$, $k$ is the smallest index of a nonterminal child node, and $m$ is the value of the counter:
\begin{align}
	\log h(S) &\triangleq \min\{ h_x^{\textbf{n}}(X) + I_{(B_k,l_k,l_{k+1})}, \log \eta_{m-1} \} \nonumber \\
	    &\hspace{5em}+ \hspace{0.25em} \rho \hspace{0.3em} + \hspace{0.2em} \max_{l_{k+2}<\hdots<l_{K+1}} \sum_{u=k+1}^{K} I_{(B_u,l_u,l_{u+1})}. \label{eq:priority2}
\end{align}
Else, if all the nonterminal child nodes of $\textbf{n}$ has children, and $m$ is the value of the counter:
\begin{equation}
	\log h(S) \triangleq h_x^{\textbf{n}}(X) + \hspace{0.2em} \rho \hspace{0.2em} + \hspace{0.2em} \log \mu_{m-1}. \label{eq:priority3}
\end{equation}
where
\begin{align*}
    \rho &\triangleq \sum_{\textbf{m}\in S\setminus\textbf{n}} \log p(r^\textbf{m} \mid x\in X, \textbf{t}), \\
    \hspace{-2.5em}h_x^{\textbf{n}}(X) &\ge \max_{x\in X}\log p(x^{\textbf{n}}) \text{ is an upper bound on the semantic prior}, \\
    \eta_m &\triangleq \parbox[t]{0.8\textwidth}{the objective function value of the $m^{\scriptsize th}$ most probable derivation trees obtained on line \ref{line:parser_branch_recursion_step},} \\
    \mu_m &\triangleq \parbox[t]{0.8\textwidth}{$m^{\scriptsize th}$ highest value of $p(A \to B_1\hspace{-1mm}\nobreak:\nobreak\hspace{-1mm}f_1 \hdots B_K\hspace{-1mm}\nobreak:\nobreak\hspace{-1mm}f_K \mid x\in X,\bm{t})$ obtained on line \ref{line:parser_branch_complete_step}.}
\end{align*}
The max in the third term of equation \ref{eq:priority2} can be computed via dynamic programming with running time $\mathcal{O}(K^2)$. In the equation for $\rho$, the sum over $\textbf{m}\in S\setminus\textbf{n}$ is over all nodes in the incomplete derivation tree of $S$, excluding $\textbf{n}$. To avoid recomputing $\rho$ every time $h$ is invoked, our implementation stores it in every search state. Its initial value is $0$. In algorithm \ref{alg:parser_branch}, on line \ref{line:parser_branch_substitute_child_subtree}, the log probability of the new search state $S*$ is equal to the sum of the log probability of the old state $S$ and the log probability of $S_m$. In line \ref{line:parser_branch_complete_derivation}, the log probability of the new search state is equal to the sum of the log probability of the old search state $S$ and $\log p(A\to B_1\hspace{-1mm}\nobreak:\nobreak\hspace{-1mm}f_1 \hdots B_K\hspace{-1mm}\nobreak:\nobreak\hspace{-1mm}f_K \mid x\in X, \bm{t})$. Our implementation then uses this quantity directly as $\rho$ in the above heuristic. The heuristic also has the nice property that when a search state is marked \texttt{COMPLETE}, its heuristic value is equal to the logarithm of the objective, aside from the prior term. Thus, when computing the objective function, such as checking the termination condition in the branch-and-bound, we only need to compute the prior term.

With a sufficiently tight upper bound on the objective, this algorithm ignores a very large number of subproblems whose upper bound is too high. Figure \ref{fig:parsing_search_tree} shows the search tree for the branch-and-bound algorithm. By ignoring sets of derivation trees with an upper bound smaller than that of the highest-scoring element in the search queue, the parser can ignore a large number of improbable logical forms and derivations. Thus, with a good upper bound, the parser can run in sublinear time with respect to the size of the theory. The parser resembles a generalized version of the Earley parsing algorithm \citep{DBLP:journals/cacm/Earley70}.

\subsection{Generating sentences} \label{sec:generating_sentences}

In contrast with parsing, given a new logical form $x_*$, natural language generation is the task of finding the unknown sentence $y_*$ and derivation tree $t_*$. A straightforward way to do this in our model is to sample $t_* \mid \bm{t}, x_*$, and simply compute $y_* = \text{yield}(t_*)$. The sampling follows the generative process directly.

{\setlength{\algomargin}{0em}
\begin{algorithm2e}
\footnotesize
\let\oldnl\nl
\newcommand{\nonl}{\renewcommand{\nl}{\let\nl\oldnl}}
\SetNlSty{}{\color{RedOrange}\sffamily}{}
\SetAlgoBlockMarkers{}{}
\SetKwProg{Fn}{function}{}{}
\SetKwIF{If}{ElseIf}{Else}{if}{ }{else if}{else }{}
\SetKw{Continue}{continue}
\SetKwFunction{FBranch}{\small branch}
\SetKwFunction{FExpand}{\small expand}
\SetKwFunction{FComplete}{\small complete}
\SetKwFor{ForEach}{for each}{do}{end}
\SetKwProg{uForEach}{for each}{ do}{}
\SetKwProg{Fn}{function}{}{}
\AlgoDisplayBlockMarkers\SetAlgoVlined
\SetAlCapNameFnt{\small}
\SetAlCapFnt{\small}
\SetNoFillComment
\DontPrintSemicolon
\SetInd{0.0em}{0.8em}
    \Fn{\FBranch{derivation tree set $S$}}{
        $L$ is an empty list \;
        $\textbf{n}$ is the root of the incomplete derivation tree of $S$ \;
        $x$ is the logical form at $\textbf{n}$ \;
        \uIf{\textbf{n} has no child nodes}{
            $A$ is the nonterminal symbol of $\textbf{n}$ \;
            \Return{$\texttt{expand(} A, x \texttt{)}$}
        }\uElseIf{$\textbf{n}$ has a nonterminal child node with no children}{
            $A\to B_1\hspace{-1mm}\nobreak:\nobreak\hspace{-1mm}f_1 \hdots B_K\hspace{-1mm}\nobreak:\nobreak\hspace{-1mm}f_K$ is the production rule at $\textbf{n}$ \;
            $\textbf{c}_k$ is the first nonterminal child node of $\textbf{n}$ with no children \;
            $B_k$ is the nonterminal symbol of $\textbf{c}_k$ \;
            $m$ is the counter of $S$ \;
            $\rho$ is the log probability of $S$ \;
            \lIf{$f_k(x)$ fails}{\Return{$\varnothing$}}
            $S_m$ is the $m^{\scriptsize th}$ most probable derivation tree with root nonterminal $B_k$ and logical form $f_k(x)$, according to equation \ref{eq:generation_objective} \label{line:generator_branch_recursion_step} \;
            \If{$S_m$ exists}{
                $S^* = S \cap S_m$ is a new derivation tree set with counter $1$, the incomplete derivation tree is identical to that of $S$ except $\textbf{c}_k$ is substituted with the incomplete derivation tree of $S_m$, and the log probability is the sum of $\rho$ and the log probability of $S_m$ \label{line:generator_branch_substitute_child_subtree} \;
                $L\texttt{.add(} S^* \texttt{)}$ \;
                $L\texttt{.add(}$ a new derivation tree set identical to $S$ except its counter is $m+1 \texttt{)}$ \;
            }
        }\Else{
            $A\to B_1\hspace{-1mm}\nobreak:\nobreak\hspace{-1mm}f_1 \hdots B_K\hspace{-1mm}\nobreak:\nobreak\hspace{-1mm}f_K$ is the production rule at $\textbf{n}$ \;
            $\rho$ is the log probability of $S$ \;
            $L\texttt{.add(}$ a new derivation tree set identical to $S$ except its log probability is the sum of $\rho$ and $p(A\to B_1\hspace{-1mm}\nobreak:\nobreak\hspace{-1mm}f_1 \hdots B_K\hspace{-1mm}\nobreak:\nobreak\hspace{-1mm}f_K \mid x, \bm{t})$, and is marked \texttt{COMPLETE} $\texttt{)}$ \label{line:generator_branch_complete_derivation} \;
        }
        \Return{$L$}
    }
    \Fn{\FExpand{nonterminal $A$, logical form $x$}}{
        $L$ is an empty list \;
        \uIf{$A$ is a preterminal}{
            \For{rules $A\to w$}{
                $S^*$ is a new derivation tree set where the incomplete derivation tree consists of a root node \textbf{n} with nonterminal $A$, logical form $x$, and child node $w$ \;
                $L\texttt{.add(} S^* \texttt{)}$ \;
            }
        }\Else{
            \For{rules $A\to B_1\hspace{-1mm}\nobreak:\nobreak\hspace{-1mm}f_1 \hdots B_K\hspace{-1mm}\nobreak:\nobreak\hspace{-1mm}f_K$}{
                $S^*$ is a new derivation tree set with counter $1$, the incomplete derivation tree consists of a root node $\textbf{n}$ with nonterminal $A$, logical form $x$, and for each child node $\textbf{c}_i$, the nonterminal is $B_i$, and logical form is $f_i(x)$ \;
                \If{$f_i(x)$ did not fail for all $i=1,\hdots,K$}{
                    $L\texttt{.add(} S^* \texttt{)}$ \;
                }
            }
        }
        \Return{$L$}
    }
    \caption{Pseudocode for \texttt{branch} and \texttt{expand} in the branch-and-bound algorithm for generating the most likely sentence(s), given a logical form, which aims to maximize equation \ref{eq:generation_objective}.}
    \label{alg:generator_branch}
\end{algorithm2e}}

However, in many situations, it is desirable to find the sentence $y_*$ and derivation $t_*$ that maximize:
\begin{align}
    p(y_*, t_* \mid x_*, \bm{x}, \bm{y}) &= \int p(y_*, t_* \mid x_*, \bm{t}) p(\bm{t} \mid \bm{x}, \bm{y}), \\
        &\approx \frac{1}{N_{\text{samples}}} \sum_{\bm{t}\sim\bm{t} \mid \bm{x},\bm{y}} p(y_*, t_* \mid x_*, \bm{t}), \\
    p(y_*, t_* \mid x_*, \bm{t}) &\approx \mathds{1}\{\text{yield}(t_*)=y_*\} \prod_{\textbf{n}\in t_*} p(r^{\textbf{n}} \mid x_*^{\textbf{n}}, \bm{t}). \label{eq:generation_objective}
\end{align}
As with parsing, we assume that $N_{\text{samples}} = 1$. This is also a discrete optimization problem, albeit simpler than parsing, and we again apply branch-and-bound. Similar to the case in parsing, each search state represents a set of derivation trees, represented by an incomplete derivation tree, except that the nodes do not have any sentence positions, since $y_*$ is not known, and there is only a single logical form rather than a set of logical forms, since $x_*$ is known. The \texttt{branch} function for generation is shown in algorithm \ref{alg:generator_branch}. The algorithm is started with a derivation tree set whose incomplete derivation tree consists of a single node labeled \textsf{S} with logical form $x_*$.

The heuristic upper bound for a search state $S$ is simply:
\begin{equation}
    \log h(S) \triangleq \sum_{\textbf{m}\in S\setminus\textbf{n}} \log p(r^\textbf{m} \mid x_*^\textbf{m}, \bm{t}) = \rho,
\end{equation}
where the sum is over all nodes in the incomplete derivation tree of $S$, excluding the root node $\textbf{n}$. Note that, just as in parsing, the algorithm keeps track of this quantity in each search state as the log probability $\rho$, and so $\log h(S) = \rho$.

Just as in parsing, in order to execute line \ref{line:generator_branch_recursion_step}, we can use the augmented branch-and-bound in algorithm \ref{alg:get_kth_best} to return the $m^{\scriptsize th}$ best derivation tree that maximizes the objective over the set of derivation trees rooted at $B_k$ with logical form $f_k(x)$. Whenever line \ref{line:parser_branch_recursion_step} is first executed for a given nonterminal $B_k$ and logical form $f_k(x)$, initialize the priority queue $Q$ in algorithm \ref{alg:get_kth_best} with: $Q\texttt{.push(} S^*, h(S^*) \texttt{)}$ where $S^*$ is the search state with an incomplete derivation tree consisting of a single node at the root with nonterminal $B_k$ and logical form $f_k(x)$. The implementation for our inside-outside sampler, branch-and-bound parser and generator is available at \href{https://github.com/asaparov/grammar}{\nolinkurl{github.com/asaparov/grammar}}.

\section{Semantic parsing experiments on \textsc{GeoQuery} and \textsc{Jobs}} \label{section:semantic_parsing_experiments}

To evaluate our parser, we use the \textsc{GeoQuery} and \textsc{Jobs} datasets \citep{DBLP:conf/aaai/ZelleM96, DBLP:conf/emnlp/TangM00}. \textsc{GeoQuery} contains 880 questions about U.S. geography. Each question is labeled with a logical form in Datalog. The dataset includes a database called \textsc{GeoBase}, which, when each logical form is executed, returns the answer to the corresponding question. The \textsc{Jobs} dataset contains 640 questions about computer-related job postings (from the USENET group \texttt{austin.jobs}). Each question is also labeled with a Datalog logical form, similar to the semantic formalism of \textsc{GeoQuery}. Most question in the two datasets are interrogative sentences, but there are some imperative sentences. Figures \ref{fig:geoquery_examples} and \ref{fig:jobs_examples} showcases some examples from each dataset, respectively. The task is semantic parsing: given each sentence, predict the logical form that represents its meaning.

\begin{figure}
    \footnotesize
    \hspace*{0.5\textwidth} \makebox[\paperwidth]{%
	\begin{tabular}{>{\raggedleft}m{0.125\textwidth} m{0.65\textwidth}} \toprule
        \textbf{Sentence} & ``How large is Alaska?'' \\
        \textbf{Logical form} & \texttt{\color{RoyalBlue}answer(A,(size(B,A),const(B,stateid(alaska))))} \\ \midrule
        \textbf{Sentence} & ``How many people lived in Austin?'' \\
        \textbf{Logical form} & \texttt{\color{RoyalBlue}answer(A,(population(B,A),const(B,cityid(austin,\_))))} \\ \midrule
        \textbf{Sentence} & ``What is the biggest city in Nebraska?'' \\
        \textbf{Logical form} & \texttt{\color{RoyalBlue}answer(A,largest(A,(city(A),} \\[-0.2em]
            & \phantom{\hspace{3em}}\texttt{\color{RoyalBlue}loc(A,B),const(B,stateid(nebraska)))))}\\ \midrule
        \textbf{Sentence} & ``Give me the cities in USA?'' \\
        \textbf{Logical form} & \texttt{\color{RoyalBlue}answer(A,(city(A),loc(A,B),const(B,countryid(usa))))} \\ \bottomrule
    \end{tabular}
    }\hspace*{-\paperwidth}
    \caption{Examples of sentences and logical form labels from \textsc{GeoQuery}.}
    \label{fig:geoquery_examples}
    \vspace{1.5em}
\end{figure}

\begin{figure}
    \footnotesize
    \hspace*{0.5\textwidth} \makebox[\paperwidth]{%
	\begin{tabular}{>{\raggedleft}m{0.125\textwidth} m{0.65\textwidth}} \toprule
        \textbf{Sentence} & ``Show me programmer jobs in Tulsa?'' \\
        \textbf{Logical form} & \texttt{\color{RoyalBlue}answer(A,(job(A),title(A,T),} \\[-0.2em]
            & \phantom{\hspace{3em}}\texttt{\color{RoyalBlue}const(T,'Programmer'),loc(A,C),const(C,'tulsa')))} \\ \midrule
        \textbf{Sentence} & ``What jobs are there with a salary of more than 50000 dollars per year?'' \\
        \textbf{Logical form} & \texttt{\color{RoyalBlue}answer(A,(job(A),salary\_greater\_than(A,50000,year)))} \\ \midrule
        \textbf{Sentence} & ``What jobs in Austin require more than 10 years of experience?'' \\
        \textbf{Logical form} & \texttt{\color{RoyalBlue}answer(A,(job(A),loc(A,P),} \\[-0.2em]
            & \phantom{\hspace{3em}}\texttt{\color{RoyalBlue}const(P,'austin'),req\_exp(A,E),const(E,10)))} \\ \midrule
        \textbf{Sentence} & ``Can I find a job making more than 40000 a year without a degree?'' \\
        \textbf{Logical form} & \texttt{\color{RoyalBlue}answer(A,(job(A),} \\[-0.2em]
            & \phantom{\hspace{3em}}\texttt{\color{RoyalBlue}salary\_greater\_than(A,40000,year),{\textbackslash}+ req\_deg(A)))} \\ \bottomrule
    \end{tabular}
    }\hspace*{-\paperwidth}
    \caption{Examples of sentences and logical form labels from \textsc{Jobs}.}
    \label{fig:jobs_examples}
    \vspace{0.9em}
\end{figure}

We created a semantic grammar for the Datalog representation of \textsc{GeoQuery} and \textsc{Jobs}, specifying the ``interior'' production rules and implementing the semantic transformation functions and their inverses.\footnote{This grammar is available at \href{https://github.com/asaparov/parser/blob/master/english.gram}{\nolinkurl{github.com/asaparov/parser/blob/master/english.gram}}.} We experiment with a simple prior for the logical forms: Let $x$ be a Datalog logical form, and $x^{a,i}$ is the $i^{\scriptsize th}$ predicate or ``function'' node in $x$ in prefix order whose smallest variable is $a$ (``smallest'' in the sense that $\color{RoyalBlue}A$ is smaller than $\color{RoyalBlue}B$ is smaller than $\color{RoyalBlue}C$ etc). For example, \texttt{\color{RoyalBlue}size(A,B)} is a predicate node whose smallest variable is \texttt{\color{RoyalBlue}A}, and \texttt{\color{RoyalBlue}most(B,C,...)} is a ``function'' node whose smallest variable is \texttt{\color{RoyalBlue}B}. The prior probability of $x$ is given by $p(x) \propto \prod_{a,i} p(x^{a,i} \mid x^{a,i-1})$ where the conditional $p(x^{a,i} \mid x^{a,i-1})$ is modeled with an HDP as in section \ref{sec:hdp_structured_model}. This HDP has height 2: the first feature function is the predicate or ``function'' symbol of the input node (e.g. \texttt{\color{RoyalBlue}size} or \texttt{\color{RoyalBlue}most}), and the second feature function is the arity and ``order'' of the arguments (e.g. \texttt{\color{RoyalBlue}size(A)} vs \texttt{\color{RoyalBlue}size(A,B)} vs \texttt{\color{RoyalBlue}size(B,A)}).

We also follow \citet{DBLP:conf/acl/WongM07,DBLP:conf/aaai/LiLS13,DBLP:conf/naacl/ZhaoH15} and experiment with type-checking, where every entity is assigned a type \change{from} a type hierarchy (e.g. \texttt{\color{RoyalBlue}alaska} has type \texttt{\color{RoyalBlue}state}, \texttt{\color{RoyalBlue}state} has supertype \texttt{\color{RoyalBlue}polity}, etc), and every predicate is assigned a functional type (e.g. \texttt{\color{RoyalBlue}population} has type $\texttt{\color{RoyalBlue}polity} \to \texttt{\color{RoyalBlue}int} \to \texttt{\color{RoyalBlue}bool}$, etc). We incorporate type-checking into the semantic prior by assigning zero probability to type-incorrect logical forms. More precisely, logical forms are distributed according to the original prior, conditioned on the fact that the logical form is type-correct. Type-checking requires the specification of a type hierarchy. Our hierarchy contains 11 types for \textsc{GeoQuery} and 12 for \textsc{Jobs}. We run experiments with and without type-checking for comparison.

Following \citet{DBLP:conf/emnlp/ZettlemoyerC07}, we use the same 600 \textsc{GeoQuery} sentences for training and an independent test set of 280 sentences. On \textsc{Jobs}, we use the same 500 sentences for training and 140 for testing. We run our parser with two setups: (1) with no domain-specific supervision, and (2) using a small domain-specific lexicon and a set of beliefs (such as the fact that Portland is a city). For each setup, we run the experiments with and without type-checking, for a total of 4 experimental setups. A given output logical form is considered correct if it is semantically equivalent to the true logical form.\footnote{The result of execution of the output logical form is identical to that of the true logical form, for any grounding knowledge base/possible world.} In these experiments, we did not use a model of morphology in the grammar. We measure the precision and recall of our method, where precision is the number of correct parses divided by the number of sentences for which our parser provided output, and recall is the number of correct parses divided by the total number of sentences in each dataset.\tomcomment{Tom: what does recall mean here? Recall of what?}\abucomment{Abu: i define recall here, but did you make this comment while reading some other part of this section? if so, do you remember where? maybe we can consider moving these definitions earlier} Our results are shown compared against many other semantic parsers in table \ref{fig:geo_jobs}. Our method is labeled \textsf{PWL-LM}. The numbers for the baselines were copied from their respective papers, and so their specified lexicons/type hierarchies may differ slightly. All code for these experiments is available at \href{https://github.com/asaparov/parser}{\nolinkurl{github.com/asaparov/parser}}.

\begin{table}
\change{
	\definecolor{fadedgreen}{rgb}{0.73,0.87,0.53}
	\footnotesize
	\hspace*{0.5\textwidth} \makebox[\paperwidth]{%
 	\begin{tabular}{ l @{\hskip 6mm} c @{\hskip 6mm} c c c @{\hskip 6mm} c c c} \toprule
		\multirow{2}{*}{\textsc{Method}} & \multirow{2}{*}{\shortstack{\textsc{Additional} \\ \textsc{Supervision}}} & \multicolumn{3}{c}{\hspace{-6mm}\textsc{GeoQuery}} & \multicolumn{3}{c}{\hspace{-2mm}\textsc{Jobs}} \\
		&  & P & R & F1 & P & R & F1 \\ \midrule
		\textsf{WASP} \citep{DBLP:conf/naacl/WongM06}				& A,B	& 87.2 & 74.8 & 80.5 &  &  &  \\
		\textsf{$\lambda$-WASP} \citep{DBLP:conf/acl/WongM07}		& A,B,F	& 92.0 & 86.6 & 89.2 &  &  &  \\
		Extended \textsf{GHKM} \citep{DBLP:conf/aaai/LiLS13}		& A,B,F	& 93.0 & 87.6 & 90.2 &  &  &  \\ \hline
		\citet{DBLP:conf/uai/ZettlemoyerC05}				& C,E,F & \textbf{96.3} & 79.3 & 87.0 & 97.3 & 79.3 & 87.4 \\
		\citet{DBLP:conf/emnlp/ZettlemoyerC07}			& C,E,F & 91.6 & 86.1 & 88.8 &  &  &  \\
		\textsf{UBL} \citep{DBLP:conf/emnlp/KwiatkowksiZGS10}		& E		& 94.1 & 85.0 & 89.3 &  &  &  \\
		\textsf{FUBL} \citep{DBLP:conf/emnlp/KwiatkowskiZGS11}		& E		& 88.6 & 88.6 & 88.6 &  &  &  \\
		\citet{DBLP:conf/emnlp/WangKZ14}			& C,E	&  & 91.1 &  &  &  &  \\
		\textsf{TISP} \citep{DBLP:conf/naacl/ZhaoH15}			& E,F	& 92.9 & 88.9 & 90.9 & 85.0 & 85.0 & 85.0 \\
		\citet{DBLP:conf/acl/RabinovichSK17}	& E,F	&  & 87.1 &  &  & \textbf{92.9} &  \\
		\textsc{Coarse2Fine} \citep{DBLP:conf/acl/LapataD18}	& E,F	&  & 88.2 &  &  &  &  \\
		\citet{DBLP:conf/acl/PlataniosPRZKGT20} & E,F,G &  & 91.4  &  &  & 91.4 &  \\
		\textsf{NQG-T5-3B} \citep{DBLP:conf/acl/ShawCPT20} & E,F,G	&  & \textbf{93.7} &  &  &  &  \\
		\rowcolor{fadedgreen} \textsf{PWL-LM} $-$ lexicon $-$ type-checking	& D		& 86.9 & 75.7 & 80.9 & 89.5 & 67.1 & 76.7 \\
		\rowcolor{fadedgreen} \textsf{PWL-LM} $+$ lexicon $-$ type-checking	& D,E	& 88.4 & 81.8 & 85.0 & 91.4 & 75.7 & 82.8 \\
		\rowcolor{fadedgreen} \textsf{PWL-LM} $-$ lexicon $+$ type-checking	& D,F	& 89.3 & 77.9 & 83.2 & 93.2 & 69.3 & 79.5 \\
		\rowcolor{fadedgreen} \textsf{PWL-LM} $+$ lexicon $+$ type-checking	& D,E,F	& 90.7 & 83.9 & 87.2 & \textbf{97.4} & 81.4 & 88.7 \\ \bottomrule
	\end{tabular}
	}\hspace*{-\paperwidth} \\[0.3em]
	\hspace*{0.5\textwidth} \hspace{1em}\hspace{1.7em}\makebox[\paperwidth]{%
	\begin{tabular}{p{67mm} p{85mm}}
		Legend for sources of additional supervision: &  \\
		\textbf{A}. Training set containing 792 examples,			& \textbf{B}. Domain-specific set of initial synchronous CFG rules, \\
		\textbf{C}. Domain-independent set of lexical templates,	& \textbf{D}. Domain-independent set of interior production rules, \\
		\textbf{E}. Domain-specific initial lexicon,				& \textbf{F}. Type-checking and type specification for entities, \\
		\textbf{G}. Pre-trained on large web corpus. \\[0.3em]
	\end{tabular}
	}\hspace*{-\paperwidth}
	\vspace{0.3em}
	\caption{Results of semantic parsing experiments on the \textsc{GeoQuery} and \textsc{Jobs} datasets \citep{DBLP:conf/conll/SaparovSM17}. Precision, recall, and F1 scores are shown. The methods in the top portion of the table were evaluated using 10-fold cross validation, whereas those in the bottom portion were evaluated with an independent test set. As a consequence, the methods evaluated using 10-fold cross validation were trained on 792 \textsc{GeoQuery} examples and tested on 88 examples for each fold (hence the additional supervision label ``A'' in the above table). In contrast, the methods evaluated using an independent test set were trained on 600 \textsc{GeoQuery} examples and tested on 280 examples. The domain-independent set of interior production rules (labeled ``D'' in the above table) is described in section \ref{sec:selecting_production_rules}. Some of the above methods use the preprocessed version of data from \citet{DBLP:conf/acl/DongL16}, where entity names and numbers in the training and test sets are replaced with typed placeholders. This provides the same additional information as a typed domain-specific lexicon.} \label{fig:geo_jobs}
}
\end{table}

\begin{figure}
	\footnotesize
	\hspace*{0.5\textwidth} \hspace{1em}\makebox[\paperwidth]{%
	\begin{tabular}{l l l}
		Logical form:	& \texttt{\color{RoyalBlue}answer(A,smallest(A,state(A)))}	& \texttt{\color{RoyalBlue}answer(A,largest(B,(state(A),population(A,B))))} \\
		Test sentence:	& ``Which state is the smallest?''			& ``Which state has the most population?'' \\
		Generated:		& ``What state is the smallest?''			& ``What is the state with the largest population?''
	\end{tabular}
	}\hspace*{-\paperwidth} \\
	\caption{Examples of sentences generated from our trained grammar on logical forms in the \textsc{GeoQuery} test set \citep{DBLP:conf/conll/SaparovSM17}. Generation is performed by computing $\arg\max_{y_*,t_*}\allowbreak p(y_*, t_* \mid x_*, \bm{t})$ as described in section \ref{sec:generating_sentences}.} \label{fig:generated}
\end{figure}

Many sentences in the test set contain tokens previously unseen in the training set. In such cases, the maximum possible recall is 88.2 and 82.3 on \textsc{GeoQuery} and \textsc{Jobs}, respectively. Therefore, we also measure the effect of adding a domain-specific lexicon, which maps semantic constants like \texttt{\color{RoyalBlue}maine} to the noun ``Maine'' for example. This lexicon is analogous to the string-matching and argument identification steps in some other semantic parsers. We constructed the lexicon manually, with an entry for every city, state, river, and mountain in \textsc{GeoQuery} (141 entries), and an entry for every city, company, position, and platform in \textsc{Jobs} (180 entries).

Aside from the lexicon and type hierarchy, the only training information is given by the set of sentences $\bm{y}$, corresponding logical forms $\bm{x}$, and the domain-independent set of interior production rules, as described in section \ref{sec:selecting_production_rules}. In our experiments, we found that the sampler converges rapidly, with only 10 passes over the data. This is largely due to our restriction of the interior production rules to a domain-independent set, which provides significant information about English syntax.

We emphasize that the addition of type-checking and a lexicon are mainly to enable a fair comparison with past approaches. As expected, their addition greatly improves parsing performance. At the time of the publication of our method \citep{DBLP:conf/conll/SaparovSM17}, we achieved state-of-the-art F1 on the \textsc{Jobs} dataset. However, even without such domain-specific supervision, the parser performs reasonably well. This is a promising indication that this parser will work effectively in the broader NLU system (described in \citet{SaparovThesis2022}), and is able to correctly parse sentences with complex and nested semantics. However, we notice a common error is the incorrect determination of scope of functions like \texttt{highest}, \texttt{shortest}, etc. This is likely due to the fact that the semantic prior does not explicitly model the scope of these functions (it assumes a uniform probability on all possible scopes). Thus, a more explicit model of scope might further improve parsing performance. We found that the semantic parsing problem is easier if the logical forms of each sentence are more similar to the syntactic structure of that sentence. In the extreme case, the logical forms would be identical to the sentences themselves, in which case parsing would be trivial. Thus, there is an inevitable balancing act in designing a semantic grammar and logical formalism for natural language, where on one hand we want the parsing problem to be as simple as possible, but on the other hand, we want the logical forms to be useful for downstream tasks, such as question-answering and reasoning, and ideally the logical forms of two distinct sentences that have the same meaning should be equivalent. These are important lessons to keep in mind when designing a domain-general semantic grammar and logical formalism.

\section{Related work}

Our grammar formalism can be related to synchronous CFGs (SCFGs) \citep{DBLP:books/lib/AhoU72}, where the semantics and syntax are generated simultaneously. However, instead of modeling the joint probability of the logical form and natural language utterance $p(x,y)$, we model the factorized probability $p(x)p(y \mid x)$, where the logical form $x$ may have its own complex prior distribution $p(x)$. Modeling each component in isolation provides a cleaner division between syntax and semantics, and one half of the model can be modified without affecting the other, and this is instrumental in larger NLU model (described in \citet{SaparovThesis2022}) since in that model, the logical form is derived from a larger theory containing background knowledge. We used a CFG in the syntactic portion of our model. \change{Note that due to the coupling with semantics, our formalism is more powerful than purely syntactic CFGs: The sets of strings generated by grammars in our formalism is strictly larger than those generated by plain CFGs. In fact, any indexed grammar can be converted into a grammar in our formalism, where the stack of indices can be interpreted as the logical form \citep{DBLP:journals/jacm/Aho68}. Linear indexed grammars (LIGs) are strictly less powerful than indexed grammars, and are weakly equivalent to combinatory categorial grammars (CCGs), head grammars, and tree-adjoining grammars \citep{DBLP:journals/mst/Vijay-ShankerW94}, which in turn are strictly more powerful than CFGs.} Richer syntactic formalisms such as CCGs \citep{DBLP:books/daglib/0012570} or head-driven phrase structure grammars (HPSGs) \citep{DBLP:conf/acl/ProudianP85} could replace the syntactic component in our framework and may provide a more uniform analysis across languages. Our model is similar to lexical functional grammar (LFG) \citep{Kaplan95}, where \textit{f}-structures are replaced with logical forms. Nothing in our model precludes incorporating syntactic information like \textit{f}-structures into the logical form, and as such, LFG is realized in our framework. Including a model of morphology in our grammar \change{furthers the comparison to} LFG. Our approach can be used to define new generative models of these grammatical formalisms. We implemented our method with a particular semantic formalism, but the grammatical model is agnostic to the choice of semantic formalism or the language. As in some previous parsers, our parsing problem can be related to the problem of finding shortest paths in hypergraphs using A* search \citep{DBLP:conf/iwpt/KleinM01,DBLP:conf/naacl/KleinM03,DBLP:conf/acl/PaulsK09,DBLP:conf/acl/PaulsKQ10,DBLP:journals/dam/GalloLP93}.

\section{Future work} \label{sec:language_module_future_work}

There is significant room for future work and exploration in the subject presented in this manuscript. In this section, we discuss shortcomings of various aspects of our approach, and give suggestions for how to overcome them.

The performance of our parser and generator depend heavily on the production rules of the grammar. Although the preterminal production rules are induced during training, we had to specify the other production rules by hand. While this does give us a great deal of control over the grammar, and enables us to incorporate prior knowledge about the English language into the grammar, it is very time-consuming. It would be valuable to look into ways in which these production rules can be induced from data. Recall that every production rule in our grammar is annotated with semantic transformation functions. These functions are intimately tied with the semantic formalism and effectively implement a theory of formal semantics. It would also be valuable to explore whether these transformation functions can be learned as well. One promising direction would be to decompose the semantic transformation functions into a sequence of elementary ``instructions.'' Each semantic transformation function could then be equivalently written as short programs in a simple programming language. We could then induce the semantic transformation functions by searching over the space of these short programs, perhaps by attempting to add or remove instructions, etc. However, it is not clear how much grammar induction would improve our current grammar for English. But such an approach would certainly help to learn grammar for other languages, about which we have much less knowledge. The statistical efficiency of our approach could greatly aid in natural language processing for low-resource languages, for which training data is very scarce.

During parsing, our method uses an upper bound on the objective function (as defined in equations \ref{eq:priority}, \ref{eq:priority2}, and \ref{eq:priority3}) that takes into account syntactic information. While this works well enough for our purposes, it may be possible to further improve the performance of the parser by defining tighter upper bound, possibly by taking into account semantic information.

Our semantic parsing model assumes that the sentences are noise-less: there are no spelling or grammatical errors in the utterances. This assumption helps to simplify the problem and to focus the scope of the thesis more onto language understanding and reasoning. But real-world language is noisy, and thus further work to extend the semantic parsing model to noisy settings is warranted. To properly handle grammatical errors, additional ``incorrect'' production rules must be added to the grammar, such as a rule where the grammatical number of the subject noun and the verb do not agree, or a rule where the subject is dropped entirely (and left to be inferred from context). Grammar induction could be used to learn these ``incorrect'' production rules. A possible way to handle spelling errors is to add another step to the generative process (as described in section \ref{sec:language_module_generative_process}). This extra step would take the correctly-spelled sentence as its input and create errors, such as insertions, deletions, or substitutions of characters. During inference, this process is inverted: Given the noisy sentence as input, the parser first needs to infer the correctly-spelled sentence (which is now latent), and then proceed with the parsing algorithm as described earlier in this chapter.

\change{The syntactic component of our grammatical formalism is a CFG, which is a \emph{projective} model of grammar. That is, in any derivation tree of a sentence, the leaves of any subtree form a contiguous substring of the sentence. For example, in the sentence ``John saw a dog which was a Yorkshire Terrier yesterday,'' the object noun phrase is ``dog which was a Yorkshire Terrier,'' which appears contiguously in the sentence. However, natural languages exhibit non-projectivity, such as in the example ``John saw a dog yesterday which was a Yorkshire Terrier,'' where the object noun phrase is now split by the adverb ``yesterday'' \citep{DBLP:conf/naacl/McDonaldPRH05}. However, techniques such as feature passing can be used to model non-projective phenomena such as syntactic movement in non-transformational models of grammar \citep{Gazdar81unboundeddependencies}. In principle it is also possible to replace the CFG with a non-projective grammar formalism such as a mildly non-projective dependency grammar \citep{DBLP:journals/coling/Kuhlmann13,Bodirsky:Kuhlmann:etal:05}.}

\subsection{Modeling context}

The logical forms in our model are assumed to be context-independent. Conditioned on the theory, they are independently and identically distributed. This assumption greatly simplifies the natural language that we need to be able to parse. While it helps to focus the scope of the thesis, it is not representative of real-world language. In real language, the distribution of a sentence is highly dependent on the sentences that precede it, even when conditioned on the theory, which contains all of the background knowledge. For example, this assumption disallows inter-sentential coreference (e.g. pronouns that can refer to objects mentioned in other sentences). Our model also assumes that the universe of discourse does not vary, and so the sentence ``All of the children are asleep'' would mean that, literally, every child in the universe is sleeping. The more likely meaning of the sentence is that all of the children within the local area, such as the home or town, are sleeping. The definite article ``the'' often indicates the uniqueness of an object: ``the tallest mountain'' indicates that there is exactly one tallest mountain. However, this is not the case in the example: ``A cat walked into the room. The cat purred.'' Here, ``the cat'' does not imply that there is exactly one cat in the universe. Rather, it means that the cat is unique in the context. The universe of discourse can change across sentences (and sometimes even within sentences). Relaxing the assumption that logical forms are context-independent would enable our parser to correctly understand these example sentences. To relax this assumption, our model must be augmented with a model of context. See \citet{SaparovThesis2022} for a more concrete proposal for a model of context.

\begin{appendices}

\section{Gibbs sampling for the Dirichlet process} \label{section:dp_gibbs}

The \emph{Chinese restaurant process} (CRP) representation of the Dirichlet process enables efficient inference using \emph{Markov chain Monte Carlo} (MCMC) methods. Suppose that we are given $\bm{y}\triangleq\{y_1,\hdots,y_n\}$ observations and we wish to infer the values of the latent variables: $\phi_i$ and $z_i$. A Gibbs sampling algorithm can be derived, where initial values for $\phi_i$ and $z_i$ are selected, $\smash{\phi_i^{(0)}}$ and $\smash{z_i^{(0)}}$, and for each iteration $t$, we sample new values of $\smash{\phi_i^{(t)}}$ and $\smash{z_i^{(t)}}$. One straightforward initialization for $\smash{\phi_i^{(0)}}$ and $\smash{z_i^{(0)}}$ is to assign each observation to its own table: $\smash{\phi_i^{(0)} = y_i}$ and $\smash{z_i^{(0)} = i}$ for $i=1,\hdots,n$. Note that the value of $\smash{\phi_i^{(t)}}$ is deterministic and equal to $\smash{y_{z_j^{(t)}}}$ for all $\smash{z_j^{(t)} = i}$. Thus, only $\smash{z_i^{(t)}}$ needs to be sampled at each iteration (for all $i=1,\hdots,n$). In Gibbs sampling, each random variable is sampled from its conditional distribution given all other variables: $\smash{z_i^{(t+1)} \sim z_i \mid \bm{\phi}^{(t)}, \bm{z}_{-i}^{(t)}, \bm{y}}$.
\begin{align}
    p(z_i \mid \bm{\phi}, \bm{z}_{-i}, \bm{y}) &\propto p(\bm{z}, \bm{\phi}, \bm{y}), \\
        &= p(z_1, \hdots, z_n) \prod_{j=1} p(\phi_j) \prod_{j=1}^n p(y_j \mid \bm{\phi}, z_j), \\
        &\propto p(z_{\pi(1)}, \hdots, z_{\pi(n)}) \mathds{1}\{y_i = \phi_{z_i}\}, \\
    \hspace{-1.5em}p(z_i = k \mid \bm{\phi}, \bm{z}_{-i}, \bm{y}) &\propto
            \begin{cases}
    			\mathds{1}\{y_i = \phi_{k}\} \frac{n_k}{\alpha + n - 1} & \text{if } n_k > 0, \\
    			\mathds{1}\{y_i = \phi_{k}\} \frac{p(\phi_k = y_i)\alpha}{\alpha + n - 1} & \text{if } n_k = 0,
    		\end{cases}
\end{align}
where $n_k$ is the number of customers sitting at table $k$ not including the $i^{\scriptsize th}$ customer, $\bm{\phi}\triangleq\{\phi_1,\phi_2,\hdots\}$ and $\bm{z}_{-i} = \bm{z} \setminus \{z_i\}$ is the set of all $z_j$ \emph{except} $z_i$, and $\mathds{1}\{\cdot\}$ is $1$ if the condition is true and zero otherwise. In this derivation, we used exchangeability to change the order of the table assignments $\bm{z}$ so that $z_i$ is the last assignment. After sufficiently many iterations, the distribution of the samples $\smash{\phi_i^{(t)}}$ and $\smash{z_i^{(t)}}$ will approach the true posterior $p(\bm{\phi},\bm{z}\mid \bm{y})$.

\section{Gibbs sampling for the hierarchical Dirichlet process} \label{section:hdp_gibbs}

The Gibbs sampling update can be derived similarly to the DP case: Given $(\bm{x}, \bm{y}, \bm{z})$, $\bm{\phi}$ and $\bm{\psi}$ can be computed deterministically. Thus we only need to sample each $z_i^{\textbf{n}}$:
\begin{align}
    \hspace{-2.6em}p(&z_i^{\textbf{n}} \mid \bm{\phi}, \bm{\psi}, \bm{z}^{-\textbf{n}}, \bm{z}_{-i}^{\textbf{n}}, \bm{x}, \bm{y}) \propto p(\bm{z}, \bm{\phi}, \bm{\psi}, \bm{x}, \bm{y}), \\
        &\hspace{4em}= \prod_{j=1} p(\phi_j) \prod_{\textbf{n}} p(z_1^{\textbf{n}}, z_2^{\textbf{n}}, \hdots) \prod_{j=1} p(\psi_j^{\textbf{n}} | \bm{\phi}, \bm{\psi}^{-\textbf{n}}, z_j^{\textbf{n}}) \prod_{j=1} p(y_j | \bm{\psi}, x_j), \\
        &\hspace{4em}\propto
            \begin{cases}
                p(z_{\pi(1)}^{\textbf{0}}, z_{\pi(2)}^{\textbf{0}}, \hdots) \mathds{1}\{\psi_i^{\textbf{0}} = \phi_{z_i^{\textbf{0}}}\} & \text{if } \textbf{n} = \textbf{0}, \\
                p(z_{\pi(1)}^{\textbf{n}}, z_{\pi(2)}^{\textbf{n}}, \hdots) \mathds{1}\{\psi_i^{\textbf{n}} = \psi_{z_i^{\textbf{n}}}^{\text{parent}(\textbf{n})}\} & \text{otherwise},
            \end{cases} \\
    \hspace{-2.3em}p(&z_i^{\textbf{n}} = k \mid \bm{\phi}, \bm{\psi}, \bm{z}^{-\textbf{n}}, \bm{z}_{-i}^{\textbf{n}}, \bm{x}, \bm{y}) \propto \\
            &\hspace{4em}\begin{cases}
    			\mathds{1}\{\psi_i^{\textbf{0}} \hspace{-0.1em}=\hspace{-0.1em} \phi_k\} \frac{n_k^{\textbf{0}}}{\alpha^{\textbf{0}} + n^{\textbf{0}}} & \text{if } \textbf{n} = \textbf{0}, n_k^{\textbf{n}} > 0, \\
    			\mathds{1}\{\psi_i^{\textbf{0}} \hspace{-0.1em}=\hspace{-0.1em} \phi_{\text{new}}\} \frac{p(\phi_{\text{new}} = \psi_i^{\textbf{0}})\alpha^{\textbf{0}}}{\alpha^{\textbf{0}} + n^{\textbf{0}}} & \text{if } \textbf{n} = \textbf{0}, n_k^{\textbf{n}} = 0, \\
    			\mathds{1}\{\psi_i^{\textbf{n}} \hspace{-0.1em}=\hspace{-0.1em} \psi_k^{\text{parent}(\textbf{n})}\} \frac{n_k^{\textbf{n}}}{\alpha^{\textbf{n}} + n^{\textbf{n}}} & \text{if } \textbf{n} \ne \textbf{0}, n_k^{\textbf{n}} > 0, \\
    			\mathds{1}\{\psi_i^{\textbf{n}} \hspace{-0.1em}=\hspace{-0.1em} \psi_{\text{new}}^{\text{parent}(\textbf{n})}\} \frac{f^{\text{parent}(\textbf{n})}(\psi_i^{\textbf{n}})\alpha^{\textbf{n}}}{\alpha^{\textbf{n}} + n^{\textbf{n}}} & \text{if } \textbf{n} \ne \textbf{0}, n_k^{\textbf{n}} = 0,
    		\end{cases}
\end{align}
where $n_k^{\textbf{n}}$ is the number of customers at node $\textbf{n}$ sitting at table $k$ \emph{not including} the customer currently being resampled, $n^{\textbf{n}}$ is the total number of customers at node $\textbf{n}$ (also not including the current customer), and $f^\textbf{m}(v)$ is shorthand for $p(\psi_{\text{new}}^{\textbf{m}} = v \mid \bm{\phi}, \bm{\psi}, \bm{z}^{-\textbf{n}}, \bm{z}_{-i}^{\textbf{n}}, \bm{x}, \bm{y})$ for any node $\textbf{m}$. Note that in the case where $\textbf{n}\ne\textbf{0}$, sampling $z_i^{\textbf{n}}$ requires computing $f^{\text{parent}(\textbf{n})}(\psi_i^{\textbf{n}})$, i.e. the probability of a customer at node $\textbf{n}$ choosing to sit a ``new'' table $p(\psi_{\text{new}}^{\text{parent}(\textbf{n})})$. This value can be computed recursively, so if the node $\textbf{m}\ne\textbf{0}$:
\begin{equation}
    f^{\textbf{m}}(v) = \frac{\alpha^{\textbf{m}} f^{\text{parent}(\textbf{m})}(v)}{\alpha^{\textbf{m}} + n^{\textbf{m}}} + \hspace{-1em}\sum_{\{k':n_{k'}^{\textbf{m}} > 0\}}\hspace{-1em} \frac{n^{\textbf{m}}_{k'} \mathds{1}\{\psi_{k'}^{\text{parent}(\textbf{m})} \hspace{-0.4em}= v\}}{\alpha^{\textbf{m}} + n^{\textbf{m}}}. \label{eq:new_table_probability_non_root}
\end{equation}
In the case where $\textbf{m}=\textbf{0}$:
\begin{equation}
    f^{\textbf{0}}(v) = \frac{\alpha^{\textbf{0}} p(\phi_{\text{new}} = v)}{\alpha^{\textbf{0}} + n^{\textbf{0}}} + \hspace{-1em}\sum_{\{k':n_{k'}^{\textbf{0}} > 0\}}\hspace{-1em} \frac{n^{\textbf{0}}_{k'} \mathds{1}\{\phi_{k'} = v\}}{\alpha^{\textbf{0}} + n^{\textbf{0}}}. \label{eq:new_table_probability_root}
\end{equation}
If $z_i^{\textbf{n}}$ is sampled to be a ``new'' table, a new customer will appear in the parent node of $\textbf{n}$, and its table assignment must be sampled next. This new customer may itself be assigned to a new table, and so this process continues recursively until a customer sits at a non-empty table, or a customer sits at an empty table at the root node $\textbf{0}$. The computation required in this recursive sampling procedure overlaps heavily with that in computing the probabilities in equations \ref{eq:new_table_probability_non_root} and \ref{eq:new_table_probability_root}, so they should be done simultaneously to avoid wasted computation.

In our code, for each iteration of Gibbs sampling, we traverse the tree nodes $\textbf{n}$ in prefix order, and resample $z_i^{\textbf{n}}$ in random order.

In many applications, including in our semantic parsing approach, we need to compute the probability of a new observation $y_{n+1}$, given its source node $x_{n+1}$ and previous observations $(\bm{x},\bm{y})$:
\begin{align}
    p(y_{n+1} \mid x_{n+1}, \bm{x}, \bm{y}) &= \int p(y_{n+1} \mid x_{n+1}, \bm{z}) p(\bm{z} \mid \bm{x}, \bm{y}) d\bm{z}, \label{eq:hdp_posterior_predictive_probability} \\
        &\approx \frac{1}{N_{\text{samples}}} \sum_{\bm{z}^{(t)} \sim \bm{z} \mid \bm{x}, \bm{y}}\hspace{-0.6em} p(y_{n+1} \mid x_{n+1}, \bm{z}^{(t)}, \bm{\phi}^{(t)}, \bm{\psi}^{(t)}).
\end{align}
The integral is approximated as a sum over posterior samples of $\bm{z}$, which can be obtained using the MCMC algorithm described above. However, we find in our experiments that the posterior is concentrated at a single point, and it suffices to keep only the final sample (i.e. $N_{\text{samples}}=1$) as a point estimate of $\bm{z}, \bm{\psi}, \bm{\phi}$. In either case, we can compute the quantity within the sum:
\begin{equation}
    p(y_{n+1} \mid x_{n+1}, \bm{z}, \bm{\phi}, \bm{\psi}) = p(\psi_{\text{new}}^{x_{n+1}} = y_{n+1} \mid \bm{z}, \bm{\phi}, \bm{\psi}).
\end{equation}
This quantity can be computed as in equations \ref{eq:new_table_probability_non_root} and \ref{eq:new_table_probability_root} (but since we are not resampling $z_i^{\textbf{n}}$, we don't exclude any customers in the $n_k^{\textbf{m}}$ terms).

The above can be extended to the case where rather than $1$ new observation, there are $k$ new observations, and we want to compute their joint probability:
\begin{equation}
    p\bigg(\bigcap_{i=1}^k y_{n+i} \hspace{0.2em}\bigg|\hspace{0.2em} \bm{x}, \bm{y}, \bigcap_{i=1}^k x_{n+i} \bigg) = \prod_{i=1}^k p\bigg(y_{n+i} \hspace{0.2em}\bigg|\hspace{0.2em} \bm{x}, \bm{y}, \bigcap_{j=1}^i x_{n+j}, \bigcap_{j=1}^{i-1} x_{n+j} \bigg). \label{eq:hdp_joint_posterior_predictive_probability}
\end{equation}
So to compute this, first compute the probability of the first observation $y_{n+1}$ alone. Next, add $(x_{n+1},y_{n+1})$ to the HDP (treat them as part of $\bm{x}$ and $\bm{y}$) and compute the probability of $y_{n+2}$ alone. Repeat until all $k$ probabilities are computed and then return the product. We observe that the joint probability does not factorize over each observation. This is due to the ``rich get richer'' effect observed in the Chinese restaurant process: If one observation is sampled, the same observation is more likely to be sampled in the future, since future customers are more likely to sit at tables with existing customers. And so the distribution is not i.i.d.

But as the number of customers $n$ becomes very large, the effect of $\alpha$ and any single customer on the distribution of the next observation becomes negligible, and so the distribution becomes more i.i.d.:
\begin{equation}
    \lim_{n\to\infty} p\bigg(\bigcap_{i=1}^k y_{n+i} \bigg| \bm{x}, \bm{y}, \bigcap_{i=1}^k x_{n+i} \bigg) = \lim_{n\to\infty} \prod_{i=1}^k p\bigg( y_{n+i} \bigg| \bm{x}, \bm{y}, \bigcap_{i=1}^k x_{n+i} \bigg).
\end{equation}
This fact can be useful when approximating
\begin{equation}
    \hspace{-0.8em}p\bigg(\bigcap_{i=1}^k y_{n+i} \hspace{0.2em}\bigg|\hspace{0.2em} \bm{x}, \bm{y}, \bigcap_{i=1}^k x_{n+i} \bigg) \approx \prod_{i=1}^k p\bigg( y_{n+i} \hspace{0.2em}\bigg|\hspace{0.2em} \bm{x}, \bm{y}, \bigcap_{i=1}^k x_{n+i} \bigg),
\end{equation}
when $n$ is large.

\subsection{Learning the concentration parameter $\alpha$}

We learn the concentration parameter from the data by placing a Gamma prior on $\alpha$:
\begin{equation}
    \alpha^{\textbf{n}} \sim \text{Gamma}(a^{\textbf{n}}, b^{\textbf{n}}).
\end{equation}
An auxiliary variable sampling method can be used to infer $\alpha$, which is described in appendix A of \citet{journals/jasa/Teh06} and section 6 of \citet{Escobar1995}. The Gibbs sampling step for $\alpha^{\textbf{n}}$ is:
\begin{align}
    s^{\textbf{n}} &\sim \text{Bernoulli}\left(\frac{n^{\textbf{n}}}{\alpha^{\textbf{n}} + n^{\textbf{n}}}\right), \\
    w^{\textbf{n}} &\sim \text{Beta}(\alpha^{\textbf{n}} + 1, n^{\textbf{n}}), \\
    \alpha^{\textbf{n}} &\sim \text{Gamma}(a^{\textbf{n}} + \max\{z_i^{\textbf{n}}\} - s^{\textbf{n}}, b^{\textbf{n}} - \log w^{\textbf{n}}).
\end{align}
Here, $\max\{z_i^{\textbf{n}}\}$ is the number of occupied tables in restaurant $\textbf{n}$. The above updates assume that each node $\textbf{n}$ has an independent $\alpha^{\textbf{n}}$. However, in many scenarios, we wish to tie the concentration parameters together to improve statistical efficiency. Suppose we constrain all the concentration parameters at each level in the hierarchy to be equal. Let $L(\textbf{n})$ be defined as the level of the node $\textbf{n}$ (i.e. $L(\textbf{0}) = 0$ and $L(\textbf{n}) = L(\text{parent}(\textbf{n})) + 1$). Let $\alpha_i$ be the concentration parameter at level $i$, and so $\alpha^{\textbf{n}} = \alpha_{L(\textbf{n})}$. Let its prior be $\alpha_i \sim \text{Gamma}(a_i,b_i)$. Then, the Gibbs sampling step for $\alpha_i$ is:
\begin{equation}
    \alpha_i \sim \text{Gamma}\left(a_i + \hspace{-1em}\sum_{\{\textbf{n}:L(\textbf{n})=i\}}\hspace{-1em} (\max\{z_i^{\textbf{n}}\} - s^{\textbf{n}}), b_i - \hspace{-1em}\sum_{\{\textbf{n}:L(\textbf{n})=i\}}\hspace{-1em} \log w^{\textbf{n}}\right).
\end{equation}
This is the approach we implement in our semantic parsing model during training. Our semantic parsing model has several HDP hierarchies, and each has its own set of hyperparameters $a$ and $b$. As an example, for one such hierarchy in our semantic parsing model (corresponding to the nonterminal $\textsf{VP}_\textsf{R}$), the hyperparameters are $a_1 = 100, a_2 = 10, b_1 = 0.1, b_2 = 1$, but the other hierarchies have similar values for their hyperparameters.

\end{appendices}

\starttwocolumn
\bibliography{language_module_standalone}

\end{document}